\documentclass[journal,compsoc]{IEEEtran}
\usepackage[ruled,linesnumbered]{algorithm2e}
\usepackage[dvipsnames]{xcolor}
\usepackage{graphicx}
\usepackage{colortbl}
\usepackage{subfigure}
\usepackage{amsmath,amssymb} 
\usepackage{multirow,hhline}
\usepackage{makecell}
\usepackage{bbding}
\usepackage{tabularx}
\usepackage{mathrsfs}
\usepackage{threeparttable} 
\usepackage{setspace} 
\usepackage{nicefrac}
\usepackage{tablefootnote}
\usepackage{algpseudocode}
\usepackage{diagbox}
\usepackage{booktabs}
\usepackage{amsthm}%
\usepackage{amsfonts} 
\usepackage{bm}
\usepackage{textcomp}%
\usepackage{subcaption}
\usepackage[pagebackref=true,breaklinks=true,colorlinks,bookmarks=false]{hyperref}
\usepackage{url}
\usepackage{tikz}
\newtheorem{theorem}{Theorem}[section]

\newtheorem{remark}[theorem]{Remark}
\newtheorem{corollary}[theorem]{Corollary}
 
\newtheorem{assumption}[theorem]{Assumption}

\hypersetup{
    colorlinks=true,
    linkcolor=blue,
    filecolor=magenta,      
    urlcolor=Cyan,
    citecolor=green,
    pdftitle={KANO},
    pdfpagemode=FullScreen,
    }

\begin{document}
\title{KANO: Kolmogorov-Arnold Neural Operator for Image Super-Resolution}

\author{Chenyu~Li,
        Danfeng~Hong,~\IEEEmembership{Senior Member,~IEEE,}
        Bing Zhang,~\IEEEmembership{Fellow,~IEEE,}
        Zhaojie Pan,
        and Jocelyn~Chanussot,~\IEEEmembership{Fellow,~IEEE}
        
\IEEEcompsocitemizethanks{
\vspace{-5pt}
\IEEEcompsocthanksitem C. Li, D. Hong, and Z. Pan are with Southeast University, Nanjing 210096, China.
\IEEEcompsocthanksitem B. Zhang is with the Aerospace Information Research Institute, Chinese Academy of Sciences, 100094 Beijing, China, and the College of Resources and Environment, University of Chinese Academy of Sciences, Beijing 100049, China, and also with Southeast University, 210096 Nanjing, China.
\IEEEcompsocthanksitem J. Chanussot is with Univ. Grenoble Alpes, Inria, CNRS, Grenoble INP, LJK, Grenoble 38000, France.
}
}

\markboth{}%
{Shell \MakeLowercase{\textit{et al.}}: Bare Advanced Demo of IEEEtran.cls for IEEE Computer Society Journals}

\IEEEtitleabstractindextext{%
\begin{abstract}
The highly nonlinear degradation process, complex physical interactions, and various sources of uncertainty render single-image Super-resolution (SR) a particularly challenging task. Existing interpretable SR approaches, whether based on prior learning or deep unfolding optimization frameworks, typically rely on black-box deep networks to model latent variables, which leaves the degradation process largely unknown and uncontrollable. Inspired by the Kolmogorov-Arnold theorem (KAT), we for the first time propose a novel interpretable operator, termed Kolmogorov-Arnold Neural Operator (KANO), with the application to image SR. KANO provides a transparent and structured representation of the latent degradation fitting process. Specifically, we employ an additive structure composed of a finite number of B-spline functions to approximate continuous spectral curves in a piecewise fashion. By learning and optimizing the shape parameters of these spline functions within defined intervals, our KANO accurately captures key spectral characteristics, such as local linear trends and the peak-valley structures at nonlinear inflection points, thereby endowing SR results with physical interpretability. Furthermore, through theoretical modeling and experimental evaluations across natural images, aerial photographs, and satellite remote sensing data, we systematically compare multilayer perceptrons (MLPs) and Kolmogorov-Arnold networks (KANs) in handling complex sequence fitting tasks. This comparative study elucidates the respective advantages and limitations of these models in characterizing intricate degradation mechanisms, offering valuable insights for the development of interpretable SR techniques.
\end{abstract}

}

\maketitle

\IEEEdisplaynontitleabstractindextext

%
\IEEEpeerreviewmaketitle

\section{Introduction}\label{sec:introduction}
\IEEEPARstart{I}{mage} super-resolution (SR) techniques mitigate hardware limitations by reconstructing high-resolution (HR) images from low-resolution (LR) counterparts with high fidelity, thereby enabling a wide range of applications. Mathematically, single-image SR (SISR) aims to infer the underlying HR image from an observed LR image, which can be formulated as a reconstruction problem involving degradation mapping. Specifically, the relationship between the LR and HR images can be expressed as:
$$\mathbf{Y}=(\mathbf{K}\otimes\mathbf{X})\downarrow_s+\mathbf{N},$$
where $\mathbf{X}\in\mathbb{R}^{H \times W \times C}$ is a HR image from an observed low-resolution image $\mathbf{Y}\in\mathbb{R}^{h \times w \times C}$.The degradation process is formulated as a combination of a blurring kernel $\mathbf{K}\in\mathbb{R}^{k \times k}$ and a downsampling operator $\downarrow_s$. In addition, the noise component $\mathcal{N}$ is typically modeled as additive white Gaussian noise (AWGN) \cite{AWGN}, characterized by its standard deviation (or noise level). 

\begin{figure}[!t]
 	  \centering
 			\includegraphics[width=0.52\textwidth]{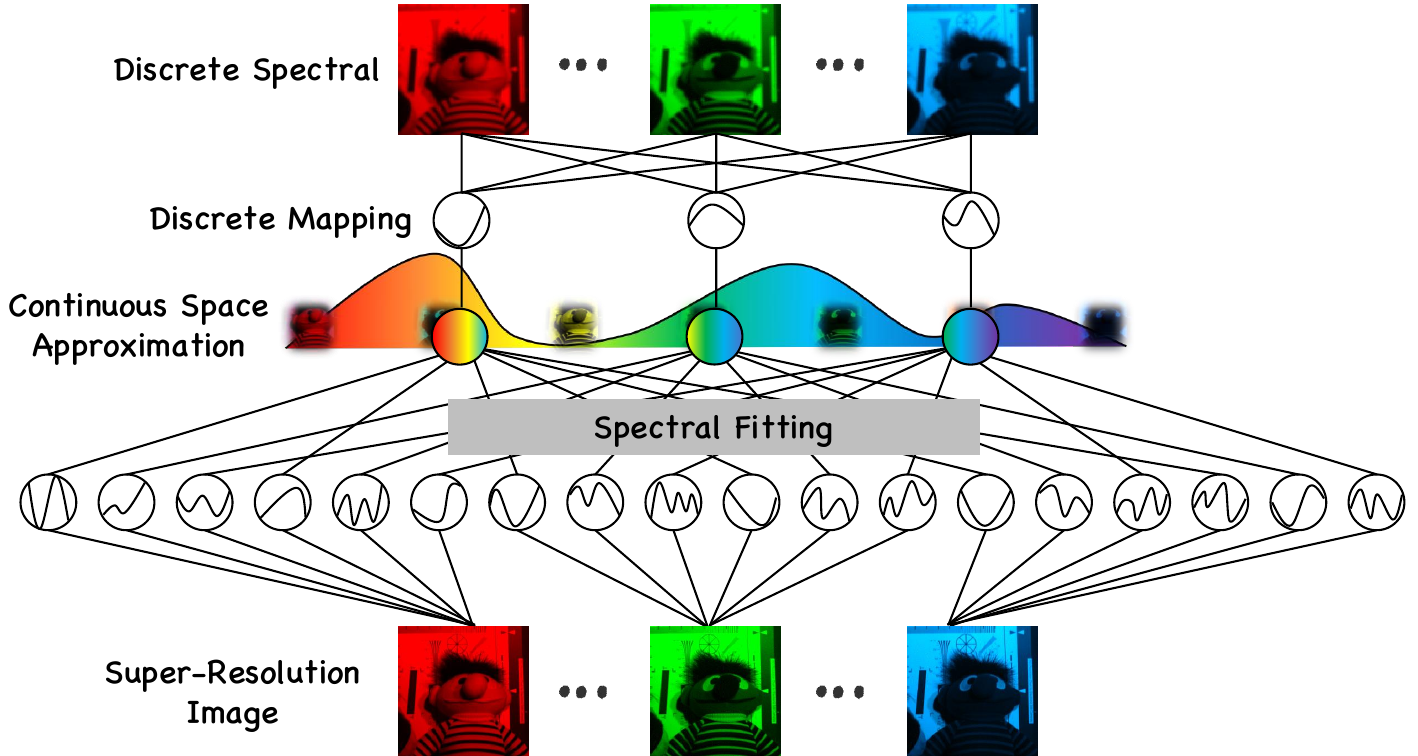}
          \caption{Illustration of the spectral fitting process using KAN. The discrete spectral measurements are transformed into an approximately continuous representation via KAN, enabling a smooth spectral profile. }
  \label{fig:KANshow}
  \end{figure}

\subsection{Evolution of Super-Resolution Techniques}
Reconstructing an HR image from its LR image counterpart is inherently challenging, as it constitutes an ill-posed inverse problem where the recovery of high-frequency details is inherently ambiguous in the absence of additional constraints. Current SR approaches are mainly aimed at learning a mapping function that restores these missing high-frequency components, thus facilitating the reconstruction of an HR image from a given observational LR image \cite{lin2018srmapping}.

Mainstream approaches to SR can be broadly categorized into two classes: traditional models with hand-crafted priors and deep learning (DL)-based techniques. Traditional SR methods typically employ manually designed priors to formulate and solve complex constrained optimization problems, leveraging principles such as sparsity and low rankness. However, the effectiveness of these methods is inherently dependent on carefully crafted prior constraints, which are often tailored to specific image characteristics and may not generalize to diverse natural images \cite{kawa2011HSI}.

Over the past decade, large-scale datasets have propelled DL-based approaches, leveraging parameterized neural networks to autonomously capture complex image priors \cite{Kim2016DRCN, lai2017DLPN}. Although achieving state-of-the-art performance in SR tasks, the black-box nature of deep networks limits the interpretability \cite{Dong2014ImageSU}. To mitigate this, recent efforts integrate DL with physical priors, bridging traditional model-based and purely data-driven paradigms \cite{Fu2022KXNet, 2024zhengjiao}. By embedding domain-specific constraints into neural architectures, these hybrid methods enhance interpretability, generalization, and robustness in SR applications.


\subsection{Approximation of Degradation Processes: Toward Interpretable Deep Learning}
Integrating physical priors into neural network design that harmonizes the strengths of expert modeling with the flexibility of data-driven DL, has attracted growing interest within the research community \cite{Fu2022KXNet}, \cite{2024zhengjiao}, which focuses primarily on the following aspects:

\noindent \textbf{Plug-and-Play (PnP).} This paradigm formulates inverse problems by integrating pre-trained denoisers to address the prior subproblem, obviating the need for hand-crafted priors \cite{K-SVD}. CNN-based denoisers have exhibited strong performance as modular components within iterative optimization schemes \cite{Deep-Denoiser-Prior}. PnP-compatible methodologies continue to advance, with SVRG \cite{ICIP} improving computational efficiency and reconstruction fidelity, while quasi-Newton methods have been incorporated through proximal denoisers \cite{PnP-Quasi-Newton-Methods}.
    
\noindent \textbf{Deep Image Prior (DIP).} Proposed by Ulyanov et al. \cite{Deep-Image-Prior}, DIP employs a deep generative network to model image priors, replacing conventional natural priors and achieving significant performance gains. \cite{PET-DIP} integrated DIP with maximum likelihood estimation, formulating a constrained optimization problem solved through ADMM. VDIP \cite{Variational-DIP} introduced hand-crafted priors on latent clear images and approximated pixel-wise distributions to mitigate suboptimality. DPL \cite{Dual-DIP} incorporated image and distortion priors to address mixed distortions. 

\noindent \textbf{Deep Unfolding $\&$ Deep Unrolling.} Deep unfolding algorithms, which reformulate the optimization process as an end-to-end trainable deep network, typically achieve superior results with fewer iterations while integrating the advantages of model-based and DL-based approaches. Several foundational works, including \cite{Model-Guided-HSISR,Spatiospectral,Deep-unfolding-proximal,Transitional-Learning} have established the theoretical and methodological underpinnings of various deep unfolding techniques.

\subsection{Challenges in Existing Interpretable Paradigm: Trade-off between Transparency and Performance}
Although IDL has garnered significant attention, its application to SR tasks remains challenging, with several critical issues yet to be addressed:

\noindent $\clubsuit$\quad  \textbf{Limit 1 (\textit{cf. \textcolor{WildStrawberry}{Simplified Hypothesis to Approximate Complex Degradation Processes.}})} Most interpretable models are founded on a specific assumption regarding the primary factor responsible for the degradation of HR images into their LR counterparts, such as noise, downsampling, or blurring. However, the actual degradation process is inherently complex, often involving the interplay of multiple factors. Such an oversimplified, single-factor assumption inevitably renders super-resolution results highly fragile and susceptible to inaccuracies.
    
\noindent $\clubsuit$\quad  \textbf{Limit 2 (\textit{cf. \textcolor{WildStrawberry}{Trade-off between Interpretability and Generalization.}})} Existing interpretability models in SR tasks primarily focus on deep approximations of local variables and the iterative process of global optimization. PnP and DIP distinctly belong to the former category, where specific neural network modules are employed to learn the physical properties of a given variable. However, this process remains inherently non-interpretable. The latter approach, epitomized by deep unfolding techniques, adheres strictly to expert-driven modeling principles. While the optimization process is highly interpretable, it fails to fully leverage the potential of deep learning in unearthing deep implicit features.

\begin{figure*}[!t]
 	  \centering
 			\includegraphics[width=1.0\textwidth]{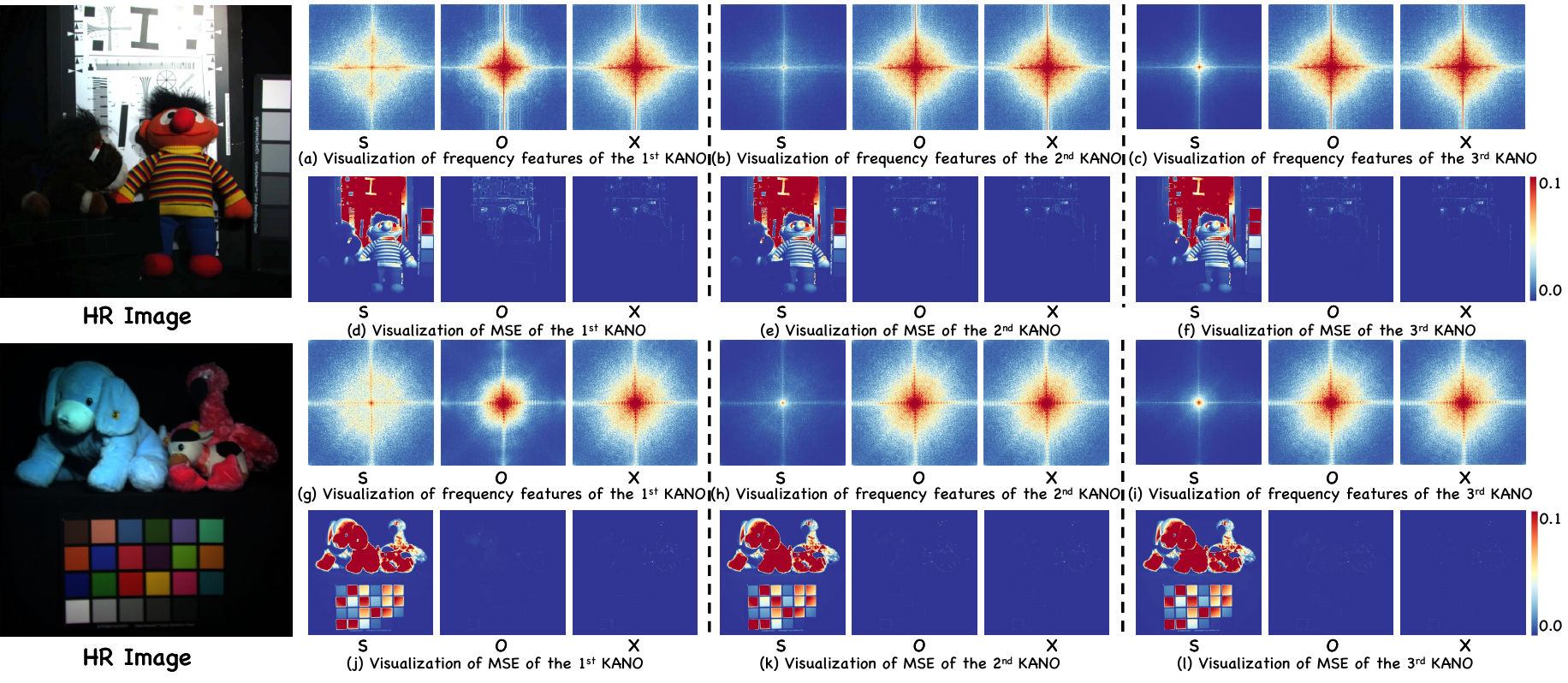}
          \caption{Visualization of $S$, $O$, and $X$ Across Iteration Steps of KANO. (a)–(c) and (g)–(i) respectively illustrate the frequency-domain characteristics of the variables $S$, $O$, and $X$. (d)–(f) and (j)–(l) present the mean squared error (MSE) between the estimated variables $S$, $O$, and $X$ and the ground truth. In early iterations, $S$ and $O$ exhibit complementary frequency features, while in later stages, the increasing similarity between $O$ and $X$ suggests a corrective effect of $S$.}
  \label{fig:medifftmse}
  \end{figure*}

\subsection{Contributions of KANO}
This work aims to develop an effective and efficient interpretable neural network to address image SR tasks while overcoming the aforementioned challenges. Specifically, our contributions are four-fold:
\begin{itemize}
\item We propose a novel interpretable operator, termed \textbf{Kolmogorov-Arnold Neural Operator (KANO)}, which, for the first time, introduces an additive structure composed of 2D and 1D spline functions to approximate the spatial degradation and continuous spectral sequence in a piecewise manner.

\item \textbf{Generalized Constraint for Degradation Kernel.} By integrating learnable activation functions with spline-based representations, the proposed approximation framework exhibits enhanced structural flexibility and inherent interpretability, enabling the modeling of spatial degradation kernels in complex imaging scenarios without relying on any prior assumptions.

\item \textbf{Decoupled approximation strategy} is introduced, wherein the HR image optimization is decomposed into: low-gradient component that enables fine-grained manipulation of global spectral trends via adjustable local shape parameters using multiple spline functions to piecewise fit (as Fig. \ref{fig:KANshow}), and a complementary residual that accounts for various perturbation-induced details.

\item \textbf{Extensive experiments conducted on diverse specialties image} substantiate the superior generalization ability of the proposed KANO. Furthermore, \textbf{the comparative analysis of MLPs and KANs} offer valuable insights for selecting and designing more efficient feature-fitting models.

\end{itemize}

The rest of this paper is organized as follows. Section \ref{sec:method} details the methodology, including the algorithm derivation and network architecture. Section \ref{sec:Experiments} showcases the experimental results and performance evaluation of KANO. The paper concludes with a future work discussion.

\section{Preliminaries}\label{Preliminaries}
\subsection{Kolmogorov–Arnold Neural (KAN)}
Kolmogorov-Arnold theorem (KAT) states that any multivariate continuous function $f$ defined on a bounded domain can be represented as a finite composition of linear combinations of continuous univariate functions \cite{gao2024kanconvergencestochasticgradientdescent,wang2024spectral-bias-kans}. Formally, the theorem can be expressed as:
\begin{equation}
\label{eq1}
f(\mathbf{x})=f(x_1,x_2,\cdots,x_n)=\sum_{q=1}^{2n+1} \mathbf{\Psi}_q \bigg(\sum\limits_{p=1}^{n} \psi_{q,p}(x_p)\bigg),
\end{equation}
where $f:[0,1]^n \to \mathbb{R}$ is a smooth function, $\psi_{q,p}:[0,1] \to \mathbb{R}$ and $\mathbf{\Psi}:\mathbb{R} \to \mathbb{R}$. 
Eq. (\ref{eq1}) can be interpreted as a two-layer nonlinear approximation network with
$2n+1$ neurons. In machine learning, its fitting capacity and generalization performance can be further enhanced by adjusting the number of layers and neurons. 

Formally, a Kolmogorov-Arnold layer with $d_{in}$ dimensional input and $d_{out}$ dimensional output can be defined as follows:
\begin{equation}
\label{eq2}
f(\mathbf{x})=\mathbf{\Psi}_{i} \circ x = [\sum\limits_{i=1}^{d_{in}} \psi_{i,1}(x_i), ...,\sum\limits_{i=1}^{d_{in}} \psi_{i,d_{out}}(x_i)],
\end{equation}
where $\Psi_{i}$ is a matrix of univariate functions. A general KAN network is a stacking of $L$ layers, given an input vector $x \in\mathbb{R}^{d_{in}}$, and the output of the general KAN network for $L$ layers can be described by
\begin{equation}
\label{eq3}
KAN(\mathbf{x})=(\mathbf{\Psi}_{L-1} \circ \mathbf{\Psi}_{L-2} \circ \cdots \circ \mathbf{\Psi}_1 \circ \mathbf{\Psi}_0)\mathbf{x}.
\end{equation}
In practice, $\psi$ is parameterized using a linear combination of the SiLU activation function and a B-spline basis,
\begin{equation}
\label{eq4}
\psi(\mathbf{x})=w_b\frac{\mathbf{x}}{1+e^{-\mathbf{x}}}+w_s\sum\limits_{i}c_iB_i(\mathbf{x}).
\end{equation}
Compared to MLPs, KAN exhibits a more flexible and controllable representational capacity. While MLPs are constrained to learning linear transformations within each layer and offer limited architectural adaptability for task-specific customization, KANs explicitly learn activation functions, enabling task-aware model tailoring and enhancing interpretability. Moreover, the degree of expressiveness in KANs can be effectively regulated by parameter adjustment, allowing precise control over the function space according to the complexity of the underlying task.

\subsection{Generalization Bound}\label{thm:generalization}
Theoretically, KAN has good generalization ability under reasonable assumptions, and its generalization error can be bounded by the following theorem:

\begin{theorem}
Given sample set $\{(x_i, y_i)\}_{i=1}^n$, if:
\begin{itemize}
    \item Input data is bounded, i.e., $\|x_i\|_2 \leq D$;
    \item Activation function space $\Psi$ satisfies Lipschitz condition, i.e., $\|\Psi(x) - \Psi(x')\|_2 \leq \rho \|x - x'\|_2$;
    \item Loss function $\mathcal{L}(f(x), y)$ is $B$-Lipschitz continuous and satisfies $\mathcal{L}(f(x), y) \in [0, M]$;
\end{itemize}
then with probability at least $1 - \epsilon$, the generalization error $R(f)$ of KAN has the following upper bound:
\begin{equation}
\label{eq5}
R(f) \leq \frac{1}{n} \sum_{i=1}^n \mathcal{L}(f(x_i), y_i) + \mathcal{O}\left( \frac{\sqrt{\log d_p} + \sqrt{\log(1/\epsilon)}}{n} \right)
\end{equation}
where $d_p$ represents the input dimension after flattening the blur kernel, and the constant term depends on network complexity (activation smoothness, weight norms, etc.).
\end{theorem}

This theorem shows that in cases with low input dimensions (such as blur kernels), KAN can maintain small generalization errors on finite samples. Experiments further verify that KAN outperforms traditional MLPs in blur kernel reconstruction tasks, showing better fitting accuracy and stronger stability.

\subsection{Activation Functions with Low-Rank Structure}
This subsection investigates the scenario in which the activation function $\Psi_l$ resides in a low-rank structure. 
\begin{theorem}
    Let \( \{(x_i, y_i)\}_{i=1}^n \) denote a set of $n$ independent samples. The loss function \( \mathcal{L}(f(x), y) \) ) is assumed to satisfy the following conditions: 
    
    \begin{assumption}\label{assumption2.2}
        The input data \( X \) is assumed to be \( \ell_2 \) bounded, i.e., \( \| X \|_2 \leq D \), where \( D > 0 \) is a predefined constant;
    \end{assumption}

    \begin{assumption}\label{assumption2.3}
    Given $v \in \mathcal{Y}$, $\mathcal{L}(\cdot, v)$ is Lipschitz in the sense that $|\mathcal{L}(u, v) - \mathcal{L}(u', v)| \leq B(v)\|u - u'\|_2 \quad \text{for any } u, u' \in \mathcal{Y}$ and $B(\cdot): \mathcal{Y} \to \mathbb{R}_{>0}$. Further suppose $\mathcal{L}(f(\cdot), \cdot) \in [0, M]$ for any $f \in \mathcal{M}$;
    \end{assumption}

    \begin{assumption}\label{assumption2.4}
    Given $v \in \mathcal{Y}$, $\mathcal{L}(\cdot, v)$ is Lipschitz in the sense that $|\mathcal{L}(u, v) - \mathcal{L}(u', v)| \leq B(v)\|u - u'\|_2 \quad \text{for any } u, u' \in \mathcal{Y}$ and $B(\cdot): \mathcal{Y} \to \mathbb{R}_{>0}$. Further suppose $\sup_{f \in \mathcal{M}} |\mathcal{L}(f(\cdot), \cdot)| \leq G(\cdot, \cdot)$ and $\mathbb{E}[G^s(x, y)] < C' < \infty \text{ for } (x, y) \sim P \text{ and some } s > 1, \quad \text{and } \mathbb{E}[B^{s'}(y_i)] < C'' < \infty \text{ for } s' > 0;$
    \end{assumption}

    \begin{assumption}\label{assumption2.5}
        Suppose $\Psi_l$ belongs to the following function space 
        \begin{equation}
        \begin{aligned}
            &\mathcal{F}_l = \{ \Psi = (\psi_1, \ldots, \psi_{d_l}) \in A_{r_l}(R_l) : \\
            &\|\Psi(x) - \Psi(x')\|_2 \leq \rho_l \|x - x'\|_2 \}
        \end{aligned}
        \end{equation}
        where $\rho_l, r_l$ and $R_l$ are some positive constants.
    \end{assumption}
\end{theorem}
Then with probability at least $1- \xi$, the generalization error $R(f)$ of KAN has the following upper bound:
\begin{equation}
    \begin{aligned}
    R(f) \leq &\frac{1}{n} \sum_{i=1}^{n} \mathcal{L}(f(\mathbf{x}_i), y_i) + \mathcal{O}\left(\frac{\sqrt{\log d_p} + \sqrt{\log \big(1 / \epsilon \big)}}{N}\right)\\
    \end{aligned}
\end{equation}
where $d_p$ represents the input dimension after flattening the blur kernel, and the constant term depends on network complexity (e.g., activation smoothness, weight norms, etc.).

\begin{theorem}\label{theorem2.1}
   \textit{Define}
\begin{equation}
    \begin{aligned}
\xi = \sum_{i=1}^{L} d_i r_i \left( \max_i B(y_i) \tilde{b} \prod_{j=i+1}^{L} \rho_j \right)^{(d_{i-1}/\nu) \vee 1},
    \end{aligned}
\end{equation}
where \( \tilde{b} = \sum_{i=1}^{L} b_i \text{ with } b_i = \tilde{C} R_i \sqrt{r_i n} \). \textit{Suppose } \( \tilde{d} := \max_i d_i > \nu \). Under assumptions \ref{assumption2.3} and \ref{assumption2.5}, we have with probability greater than  \( 1 - \epsilon \),
\begin{equation}
    \begin{aligned}
R(f) \leq &\frac{1}{n} \sum_{i=1}^{n} \mathcal{L}(f(\mathbf{x}_i), y_i) + \frac{6 \tilde{C}' (\xi)^{\nu / \tilde{d}}}{n^{(\nu / \tilde{d} + 1)/2} (\tilde{d} / \nu - 1)^{\nu / \tilde{d}}}\\
&+ \sqrt{ \frac{4M^2 \log(2 / \epsilon)}{n} } + \frac{32M \log(2 / \epsilon)}{3n}\\
    \end{aligned}
\end{equation}
\textit{for any } \( f \in \mathcal{M} \) \textit{and some constant } \( \tilde{C}' > 0 \). 
\end{theorem}
Theorem \ref{theorem2.1} generalizes the results for low-rank kernel ridge regression to a multi-layer network structure induced by KANs with activation functions belonging to a Reproducing Kernel Hilbert Space (RKHS). Besides,
\begin{remark}\label{remark2.7}
    Given a set of (fixed) activation functions $\Phi_l \in \mathcal{N}^{\otimes d_l}_K$ for $1 \leq l \leq L$, we define the function class 
    \begin{equation}
        \begin{aligned}
         \label{eq10}
            &\widetilde{\mathcal{F}}_l = \{ \widetilde{\Psi} = \Phi_l + \Psi : \Psi \in A_{r_l}(R_l), \\
            & \|\widetilde{\Psi}(x) - \widetilde{\Psi}(x')\|_2 \leq \rho_l \|x - x'\|_2\} \\
        \end{aligned}
    \end{equation}   
\end{remark}
Then, the conclusion in Theorem 4 remains true when the activation function in layer $l$ belongs to $\widetilde{\mathcal{F}}_l$, Let
\begin{equation}
\label{eq11}
    \xi_0 \equiv \sum_{i=1}^L d_i r_i \left( \left( \frac{n C''}{\tau} \right)^{2/s'} \widetilde{b} \prod_{j=i+1}^L \rho_j\right)^{(d_{i-1}/v) \vee 1} ,
\end{equation}
where $\widetilde{b} = \sum_{i=1}^L b_i$ with $b_i = \widetilde{C} R_i \sqrt{r_i n}$.

\begin{theorem}\label{theorem2.8}
Under Assumptions \ref{assumption2.4} and \ref{assumption2.5}, we have with probability greater than $1 - \epsilon - \tau - \eta$,
\begin{equation}
    \begin{aligned}
        &R(f) \leq \frac{1}{n} \sum_{i=1}^n \mathcal{L}(f(x_i), y_i) + \frac{6 \widetilde{C}'(\xi_0)^{\nu/\widetilde{d}}}{n^{(\nu/\widetilde{d}+1)/2}(\widetilde{d}/\nu-1)^{\nu/\widetilde{d}}} \\
&+ \frac{2 \sqrt{\log(2/\epsilon)}}{n^{(s-1)/(2s)}} + \frac{32 \log(2/\epsilon)}{3n^{(2s-1)/(2s)}} + \frac{2C'}{\eta n^{(s-1)/(2s)}}\\
    \end{aligned}
\end{equation}

\textit{for any $f \in \mathcal{M}$ and $\epsilon, \tau, \eta, \widetilde{C}' > 0$.}
\end{theorem}

\begin{corollary}
    Let $\hat{f} \in \mathcal{M}$ satisfy that $\sum_{i=1}^n \mathcal{L}(\hat{f}(x_i), y_i) \leq \sum_{i=1}^n \mathcal{L}(f^*(x_i), y_i)$. Suppose assumptions \ref{assumption2.2}, \ref{assumption2.2} and \ref{assumption2.4} hold, where $s \geq 2$ in assumption \ref{assumption2.4}. We have with probability greater than $1 - \epsilon - \tau - 2\eta$,
    \begin{equation}
    \begin{aligned}
        &R(\hat{f}) - R(f^*) \leq \frac{6 \widetilde{C}' (\xi_0)^{\nu/\widetilde{d}}}{n^{(\nu/\widetilde{d} + 1)/2} (\widetilde{d}/\nu - 1)^{\nu/\widetilde{d}}} 
(1 + \eta^{-1/2})\\
&\sqrt{ \frac{2(C')^{2/s} \log(2/\epsilon)}{n} }
+ \frac{32 \log(2/\epsilon)}{3n^{(2s-1)/(2s)}} + \frac{2C'}{\eta n^{(s-1)/(2s)}},\\
    \end{aligned}
    \end{equation}
for $\epsilon, \tau, \eta > 0$, where $\xi_0$ is defined in Eq. (\ref{eq11}).
\end{corollary}

Combining the arguments from Theorems \label{theorem2.8}, we can derive a generalization bound for KANs in the more general case. This applies when the activation functions of certain layers, while the activation functions of other layers have a low-rank structure, e.g., as specified in Remark \ref{remark2.7}.

\begin{figure*}[!t]
 	  \centering
 			\includegraphics[width=1.0\textwidth]{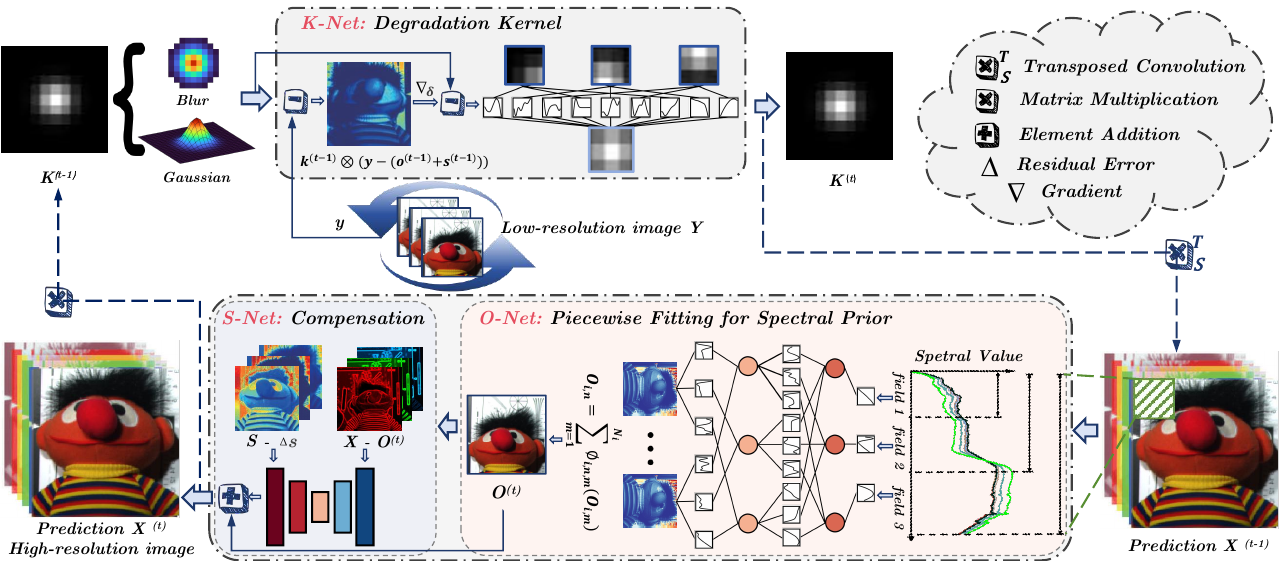}
          \caption{The proposed KANO network is formulated within an iterative optimization framework and comprises three task-specific subnetworks: (a) \textbf{K-Net} estimates the spatial degradation kernel by superimposing multiple two-dimensional spline functions, providing an interpretable representation of spatial blur. (b) \textbf{O-Net} integrates both 2D and 1D spline functions to align 2D spatial plane along the channel combined and piecewise approximate the spectral curves for per-pixel. (c) \textbf{S-Net} enforces multi-scale discrepancy compensation for the variable $\mathbf{O}$, enhancing the robustness and generalization capability under various degradation conditions.}
  \label{fig:network}
  \end{figure*}

\section{Kolmogorov-Arnold Neural Operator (KANO) in Image Super-Resolution Task}\label{sec:method}
KANO integrates optimization modeling principles with deep unfolding architectures by formulating the imaging degradation process, enabling seamless integration of physical fidelity and deep feature representation. More importantly, each variable is approximated via a tailor-designed neural network architecture specifically aligned with the domain-specific priors of the corresponding variable. The overall structure is shown in Fig. \ref{fig:network}.

\subsection{Motivation}
From the comprehensive analysis in Section \ref{Preliminaries}, we derive the following two key conclusions that serve as the foundation for the subsequent methodological developments: 
\begin{itemize}
    \item Compared to conventional MLP-based architectures, the KAN approach has superior generalization capabilities;
    \item The explicit learning of activation functions within the KAN framework not only enhances model interpretability but also provides greater flexibility in controlling the intermediate fitting process.
\end{itemize}
Typically, the solution involves jointly addressing the following two optimization subproblems:

\quad\noindent \textcolor{cyan}{\textbf{Subproblem 1. Degradation Kernel}} can essentially be modeled as a set of complex transformation matrices, whose intricate characteristics are difficult to constrain using a single regularization strategy. Moreover, the inherent "black-box" nature of deep learning models impedes the interpretability of the intermediate degradation process. In contrast, KANs, owing to their strong generalization capacity and structural simplicity, demonstrate promising capabilities in approximating a wide range of complex mapping relationships and interpretability.

\quad\noindent \textcolor{cyan}{\textbf{Subproblem 2. High-Resolution Image}} estimation is commonly constrained by either a single deep network or handcrafted regularization, which, however, often suffers from significant estimation errors due to the inherent heterogeneity between the intrinsic properties of pixels. 

To address this limitation, we propose decomposing the HR image $\mathbf{X}$ into two components $\big (\mathbf{O} + \mathbf{S}\big)$: the spectrally pure component $\mathbf{O}$ that focuses on preserving the intrinsic spectral continuity of high-dimensional data, emphasizing material-dependent spectral regularities and fidelity. In contrast, perturbation-dominated component $\mathbf{S}$ is designed to model and compensate for information degradation caused by various perturbations, thereby enhancing the robustness and integrity of the super-resolution images.

\subsection{Model Formulation and Optimization}
Giving a LR image $\mathbf{Y}\in\mathbb{R}^{h\times w \times C}$, our objective is to simultaneously recover the unknown blur kernel $\mathbf{K}\in \mathbb{R}^{k \times k}$ and the HR image $\mathbf{X}\in \mathbb{R}^{H \times W \times C}$ from the observed data. The degradation model follows a physically interpretable process:
\begin{equation}
    \begin{aligned}
     \label{eq14}
        \mathbf{Y} = \mathbf{K} \otimes \big (\mathbf{O} + \mathbf{S}\big),
    \end{aligned}
\end{equation}
where $\otimes$ denotes two-dimensional (2D) convolution and the standard downsampling operation for $\mathbf{X} = \mathbf{O} + \mathbf{S}$. The corresponding optimization problem is formulated as:
\begin{equation}
    \begin{aligned}
    \label{eq15}
        &\min\limits_{\mathbf{K},\mathbf{O}, \mathbf{S}} \frac{1}{2}\left\Vert\mathbf{Y} - \mathbf{K} \otimes \big(\mathbf{O} + \mathbf{S}\big) \right\Vert_F^2\\
        &~~~~~~~~~~~~+ \lambda_1R_1(\mathbf{K}) + \lambda_2R_2(\mathbf{O}) + \lambda_3R_3(\mathbf{S}), \\
        &\ s.t. \ \mathbf{K}_j\geq0,\ \sum_j\mathbf{K}_j=1,\forall j, \\
    \end{aligned}
\end{equation}
where the data fidelity term $\frac{1}{2}\left\Vert\mathbf{Y} - \mathbf{K} \otimes \big(\mathbf{O} + \mathbf{S}\big) \right\Vert_F^2$ encapsulates the underlying physical generation mechanism, offering explicit guidance during the iterative updates. In contrast, the prior terms \(R_1(\mathbf{K})\), \(R_2(\mathbf{O})\), and \(R_3(\mathbf{S})\) serve to regularize this inherently ill-posed problem by enforcing desirable structural properties on the solution. \(\lambda_1, \lambda_2, \lambda_3\) are trade-off regularization parameters. 

The overall optimization framework adheres to the Alternating Direction Method of Multipliers (ADMM), wherein the updates of the three sub-variables $\mathbf{K}$, $\mathbf{O}$, and $\mathbf{S}$ are approximated by proximal gradient operators. This formulation enables each customized deep network module to correspond explicitly to an iteration step of the underlying optimization algorithm, thereby implicitly guiding the optimization trajectory through structural priors.

\textbf{Degradation Kernel} is optimized by formulating a second-order (quadratic) approximation of Eq. (\ref{eq15}) with respect to $\mathbf{K}$, leading to the following closed-form expression:

\begin{equation}
\begin{aligned}\label{eq16}
&\min\limits_{\mathbf{K}}\frac{1}{2}\left\Vert\mathbf{K}-\left(\mathbf{K}^{(t-1)}-\gamma_1\nabla_k f(\mathbf{K}^{(t-1)})\right)\right\Vert_F^2 + \lambda_1\gamma_1R_1(\mathbf{K}), \\
\end{aligned}
\end{equation}
where $\gamma_1$ denotes the stepsize parameter, and $f(\mathbf{K}^{(t-1)})=\frac{1}{2}\left\Vert\mathbf{Y} - \mathbf{K}^{(t-1)} \otimes \big(\mathbf{O}^{(t-1)} + \mathbf{S}^{(t-1)} \big) \right\Vert_F^2$. According to the proximal operator $Prox(\cdot)$, for a general regularization term, Eq. (\ref{eq16}) can be expressed as:

\begin{equation} 
    \begin{aligned}
    \label{eq17}
        \mathbf{K}^{(t)}= Prox_{\lambda_1\gamma_1}\left(\mathbf{K}^{(t-1)}-\gamma_1\nabla_k f\left(\mathbf{K}^{(t-1)}\right)\right), \\
    \end{aligned}
\end{equation}
where  
\begin{equation}
    \begin{aligned}
    \label{eq18}
    &\nabla_k f(\mathbf{K}^{(t-1)}) = \mathbb{VEC}^{-1}\left({\mathbf{O}^{(t-1)} + \mathbf{S}^{(t-1)}}\right)^{T}\\
    &~~~~~~~~~~~~~~~\qquad\mathbb{VEC}\left(\mathbf{Y}-\mathbf{K}^{(t-1)} \otimes \left(\mathbf{O}^{(t-1)} + \mathbf{S}^{(t-1)} \right)\right),
    \end{aligned}
\end{equation}
where $\mathbb{VEC}(\cdot)$ means matrix vectorization and $\mathbb{VEC}^{-1}(\cdot)$ is the reverse vectorization. $(\cdot)^{T}$ is the transpose operation. 

\textbf{HR image $\mathbf{X}$} includes spectral fitting $\mathbf{O}$ and perturbation compensation $\mathbf{S}$:
\begin{itemize}
    \item The quadratic approximation of the problem in Eq. (\ref{eq15}) with respect to $\mathbf{O}$ can be derived as:
\begin{equation} 
    \begin{aligned}
    \label{eq19}
        \mathbf{O}^{(t)}= Prox_{\lambda_2\gamma_2}\left(\mathbf{O}^{(t-1)}-\gamma_2\nabla_{o} h\left(\mathbf{O}^{(t-1)}\right)\right), \\
    \end{aligned}
\end{equation}
where
\begin{equation}
    \begin{aligned}
    \label{eq20}
    &\nabla_{o} h(\mathbf{O}^{(t-1)}) \\
    &= \mathbf{K}^{(t)}\otimes_{s}^T\left(\mathbf{Y}-\mathbf{K}^{(t)} \otimes \left(\mathbf{O}^{(t-1)} + \mathbf{S}^{(t-1)} \right)\right),
    \end{aligned}
\end{equation}
where $\otimes_s^T$ is the transposed convolution operation.

    \item Similarly, $\mathbf{S}$ can be derived as:
\begin{equation} 
    \begin{aligned}
    \label{eq21}
        \mathbf{S}^{(t)}= Prox_{\lambda_3\gamma_3}\left(\mathbf{S}^{(t-1)}-\gamma_3\nabla_{s} h\left(\mathbf{S}^{(t-1)}\right)\right), \\
    \end{aligned}
\end{equation}
and
\begin{equation}
    \begin{aligned}
    \label{eq22}
    &\nabla_{s} h(\mathbf{S}^{(t-1)}) \\
    &= \mathbf{K}^{(t)}\otimes_{s}^T\left(\mathbf{Y}-\mathbf{K}^{(t)} \otimes \left(\mathbf{O}^{(t)} + \mathbf{S}^{(t-1)} \right)\right).
    \end{aligned}
\end{equation}
\end{itemize}

In contrast to current approaches that impose a single regularization constraint directly on $\mathbf{X}$, the proposed method decomposes the HR image into $\mathbf{O} + \mathbf{S}$, enabling the imposition of distinct, attribute-specific constraints in both spectral function and spatial structural domains, which facilitates a more comprehensive representation and approximation. 

\subsection{Network Design}\label{Network}
Unlike most existing blind SR approaches that primarily rely on end-to-end mappings via pure deep networks and PnP, DIP, etc., in proposed KANO, each prior network tailored for variables lacking closed-form solutions is specific to properties or constraints. 

\textcolor{cyan}{\textit{{$\blacklozenge$} $\mathcal{K}$-Net}} is derived in Theorem \ref{thm:generalization} that demonstrate the strong generalization ability and structural parsimony of the KAN architecture. According to Eq.~\ref{eq16}, K-Net can be succinctly formulated as:
\begin{equation}
    \begin{aligned}\label{eq23}
    &\mathbf{K}^{(t)}=\mathcal{K}\mathbf{\mathit{Net}}\Big(\mathbf{K}^{(t-1)}-\gamma_1\bigtriangleup\mathbf{K}^{(t-1)}\Big),
    \end{aligned}
\end{equation}
where $\bigtriangleup\mathbf{K}^{(t-1)} =\mathbb{VEC}^{-1}\left({\mathbf{O}^{(t-1)} + \mathbf{S}^{(t-1)}}\right)^{T} \mathbb{VEC}(\mathbf{Y}-\mathbf{K}^{(t-1)}$
$\otimes (\mathbf{O}^{(t-1)} + \mathbf{S}^{(t-1)}))$. 

\textcolor{cyan}{\textit{{$\blacklozenge$} $\mathcal{O}$-Net}} adopts a hybrid architecture that integrates 1D$\_$KAN and 2D$\_$KAN: four sets of two-dimensional spline functions are employed to finely align in the spatial domain. Subsequently, piecewise fitting of the spectral curves is performed with the one-dimensional spline functions. This network is built as:
\begin{equation}
    \begin{aligned}\label{eq24}
    &\mathbf{O}^{(t)}=\mathcal{O}\mathbf{\mathit{Net}}\Big(\mathbf{O}^{(t-1)}-\gamma_2\bigtriangleup\mathbf{O}^{(t-1)}\Big),
    \end{aligned}
\end{equation}
where $\bigtriangleup\mathbf{O}^{(t-1)} =\mathbf{K}^{(t)}\otimes_{s}^T(\mathbf{Y}-\mathbf{K}^{(t)} \otimes (\mathbf{O}^{(t-1)} + \mathbf{S}^{(t-1)} ))$. In addition, the 2D$\_$KAN architecture expects an input matrix with the shape $\mathbf{O}'_{i} \in \mathbb{R}^{C, H, W}$, while the 1D$\_$KAN variant operates on inputs formatted as $\mathbf{O}'_{i} \in \mathbb{R}^{H\times W, C}$.

\textcolor{cyan}{\textit{{$\blacklozenge$} $\mathcal{S}$-Net}} is to compensate for the residual information that is difficult to capture through spectral modeling, especially concentrated in edges, high frequencies, textures, and other regions. The network is built as:
\begin{equation}
    \begin{aligned}\label{eq25}
    &\mathbf{S}^{(t)}=\mathcal{S}\mathbf{\mathit{Net}}\Big(\mathbf{S}^{(t-1)}-\gamma_3\bigtriangleup\mathbf{S}^{(t-1)}\Big),
    \end{aligned}
\end{equation}
where $\bigtriangleup\mathbf{S}^{(t-1)} =\mathbf{K}^{(t)}\otimes_{s}^T\left(\mathbf{Y}-\mathbf{K}^{(t)} \otimes \left(\mathbf{O}^{(t)} + \mathbf{S}^{(t-1)} \right)\right).$ Built upon the classical U-shape architecture, S-Net leverages deeper nonlinear representations across multi-scales, enabling more comprehensive information compensation.

\subsection{Initialization and Loss Function}
\label{Loss}
Before the training, we need to initialize some parameters:
\begin{subequations}
\label{eq:init}
\begin{align}
\mathbf{X}^{(0)}  = Bicubic_s\left(\mathbf{Y}\right) \\
\mathbf{O}^{(0)}  = \mathbf{X}^{(0)} \\
\mathbf{S}^{(0)}  = \mathbf{X}^{(0)} - \mathbf{O}^{(0)} \\
\mathbf{K}^{(0)}  = Gauss_{(k_h\times1),1.0} \times Gauss_{(1\times k_h),1.0}
\end{align}
\end{subequations}
where, $Bicubic_s(\cdot)$ denotes bicubic spline interpolation, $Gauss_{(k_h \times 1),1.0}$ indicates the generation of a one-dimensional Gaussian kernel of length $k_h$ with a standard deviation of 1.0.

\begin{table*}[!t]
    \centering
\caption{A quantitative comparison of average PSNR (dB) among different SR methods, evaluated on four widely \textit{\textcolor{WildStrawberry}{Natural Images}} (Set5, Set14, BSD100 and Urban100), with three noise factors $\mathcal{N} \in \{0, 5, 15 \}$ and three sampling scale $\gamma \in \{2, 3, 4\}$. \textcolor{RubineRed}{\textbf{Bold}} represents the best result and \textcolor{cyan}{\underline{underline}} represents the second best.}
\scalebox{0.9}{
    \begin{tabular}{c|c|ccc|ccc|ccc|ccc}
        \Xhline{1.0pt} 
         \cellcolor{gray!10}{}&\cellcolor{gray!10}{\textbf{dataset}} & \multicolumn{3}{c|}{\cellcolor{gray!10}{\textbf{Set5}}} & \multicolumn{3}{c|}{\cellcolor{gray!10}{\textbf{Set14}}} & \multicolumn{3}{c|}{\cellcolor{gray!10}{\textbf{BSD100}}} & \multicolumn{3}{c}{\cellcolor{gray!10}{\textbf{Urban100}}}\\
          \hhline{>{\arrayrulecolor{gray!10}}->{\arrayrulecolor{black}}|-|-|-|-|-|-|-|-|-|-|-|-|-|}
             \multirow{-2}{*}{\textbf{\cellcolor{gray!10}{\textbf{Method}}}} & \multicolumn{1}{c|}{\cellcolor{gray!10} {\textbf{$\mathcal{N}=0$}}} & \cellcolor{gray!10}{\textbf{2}} & \cellcolor{gray!10}{\textbf{3}} &\cellcolor{gray!10}{\textbf{4}} &\cellcolor{gray!10}{\textbf{2}} &\cellcolor{gray!10}{\textbf{3}} & \cellcolor{gray!10}{\textbf{4}} & \cellcolor{gray!10}{\textbf{2}} &\cellcolor{gray!10}{\textbf{3}} &\cellcolor{gray!10}{\textbf{4}} &\cellcolor{gray!10}{\textbf{2}} & \cellcolor{gray!10}{\textbf{3}} & \cellcolor{gray!10}{\textbf{4}}  \\
        \hline
        \hline
        \multicolumn{2}{c|}{EDSR (CVPR 2017)\cite{EDSR}} & 33.99& 31.57& 30.56& 30.63& 28.50 & 27.64& 29.80& 27.75& 26.90& 27.54& 25.59& 24.84\\
        \multicolumn{2}{c|}{IMDN (ACMM 2019)\cite{IMDN}}  & 33.43& 31.38& 30.10& 30.18& 28.28 & 27.24& 29.41& 27.59& 26.64& 27.06& 25.29& 24.29 \\
        \multicolumn{2}{c|}{SwinIR (ICCV 2021)\cite{swinir}}  & 34.26& 31.87& 30.52&  \textcolor{cyan}{\underline{31.24}}& 28.78 & 27.70& 30.18& 27.92& 26.91& \textcolor{cyan}{\underline{28.64}}& 26.01& 25.05\\
        \multicolumn{2}{c|}{ELAN (ECCV 2022)\cite{ELAN}}  & {{34.33}}& 31.64&30.40&30.87& 28.64& 27.59& 30.01& 27.87& 26.91& 27.79& 25.89& 25.03\\
        \multicolumn{2}{c|}{KXNet (ECCV 2022)\cite{kxnet}} & 34.10& \textcolor{cyan}{\underline{32.34}}&  \textcolor{cyan}{\underline{31.00}}& 30.77& 28.87 & \textcolor{cyan}{\underline{27.85}}& 29.91& 28.01& \textcolor{cyan}{\underline{27.03}}& 27.89 & 26.11& 25.13\\
        \multicolumn{2}{c|}{HAT (CVPR 2023)\cite{HAT}} & 34.19& 31.95& 30.36& 31.16& \textcolor{cyan}{\underline{28.95}} & 27.67& \textcolor{cyan}{\underline{30.10}}& \textcolor{cyan}{\underline{28.03}}& 26.84& 28.45 & \textcolor{RubineRed}{\textbf{26.88}}& \textcolor{cyan}{\underline{25.15}}\\
        \multicolumn{2}{c|}{SRFormer (ICCV 2023)\cite{srformer}}  & 33.85& 31.61& 30.15& 30.63& 28.66 & 27.54& 29.71& 27.82& 26.74& 27.54& 25.81& 24.79\\
        \multicolumn{2}{c|}{IPG (CVPR 2024)\cite{IPG}}  & 32.98& 31.48& 29.85& 30.62& 28.52& 27.11& 29.39& 27.67& 26.59& 28.31& 26.14& 24.66\\
        \multicolumn{2}{c|}{CFSR (TIP 2024)\cite{CFSR}}  & \textcolor{cyan}{\underline{34.65}}& 31.46& 30.25& 31.13& 28.44 & 27.38& 30.22& 27.70& 26.71& 28.01 & 25.47& 24.48\\
        \multicolumn{2}{c|}{CRAFT (TPAMI 2025)\cite{CRAFT}}  & 33.44& 31.33& 29.98& 30.88& 28.41 & 27.22& 29.72& 27.66& 26.63& 28.20 & 25.73& 24.68 \\
        \hline
        \multicolumn{2}{c|}{\cellcolor{red!5} {\textcolor{RubineRed}{\textbf{KANO (Ours)}}}} & \cellcolor{red!5} {\textcolor{RubineRed}{\textbf{34.92}}}& \cellcolor{red!5} {\textcolor{RubineRed}{\textbf{32.39}}}& \cellcolor{red!5} {\textcolor{RubineRed}{\textbf{31.28}}}& \cellcolor{red!5} {\textcolor{RubineRed}{\textbf{31.38}}}& \cellcolor{red!5} {\textcolor{RubineRed}{\textbf{28.96}}} & \cellcolor{red!5} {\textcolor{RubineRed}{\textbf{28.03}}}& \cellcolor{red!5} {\textcolor{RubineRed}{\textbf{30.35}}}& \cellcolor{red!5} {\textcolor{RubineRed}{\textbf{28.10}}}& \cellcolor{red!5} {\textcolor{RubineRed}{\textbf{27.15}}}& \cellcolor{red!5} {\textcolor{RubineRed}{\textbf{28.67}}} & \cellcolor{red!5} {\textcolor{cyan}{\underline{26.42}}}& \cellcolor{red!5} {\textcolor{RubineRed}{\textbf{25.54}}}\\
        \hline
        \hline
        \multicolumn{2}{c|}{\cellcolor{gray!10} {\textbf{$\mathcal{N}=5$}}} & \cellcolor{gray!10}{\textbf{2}} & \cellcolor{gray!10}{\textbf{3}} &\cellcolor{gray!10}{\textbf{4}} &\cellcolor{gray!10}{\textbf{2}} &\cellcolor{gray!10}{\textbf{3}} & \cellcolor{gray!10}{\textbf{4}} & \cellcolor{gray!10}{\textbf{2}} &\cellcolor{gray!10}{\textbf{3}} &\cellcolor{gray!10}{\textbf{4}} &\cellcolor{gray!10}{\textbf{2}} & \cellcolor{gray!10}{\textbf{3}} & \cellcolor{gray!10}{\textbf{4}}  \\
        \hline
        \hline
        \multicolumn{2}{c|}{EDSR (CVPR 2017)\cite{EDSR}} & 31.69& 30.20& 29.32& 28.86& 27.55 & {{26.80}}& 28.07& 26.90& 26.17& 26.28& 24.90& 24.25\\
        \multicolumn{2}{c|}{IMDN (ACMM 2019)\cite{IMDN}}  & 31.55& 30.16& 29.02& 28.76& 27.51 & 26.59& 28.00& 26.87& 26.05& 26.07& 24.77& 23.86 \\
        \multicolumn{2}{c|}{SwinIR (ICCV 2021)\cite{swinir}}  & 31.90& 30.48& 29.35&  \textcolor{cyan}{\underline{29.33}}& 27.79 & {{26.88}}& \textcolor{cyan}{\underline{28.34}}& 27.05& 26.22& \textcolor{cyan}{\underline{27.11}}& 25.29& 24.43\\
        \multicolumn{2}{c|}{ELAN (ECCV 2022)\cite{ELAN}}  & 31.94& 30.30& 29.39&29.12& 27.74& {{26.87}}& {{28.24}}&27.03& 26.23& 26.60& 25.24& 24.48\\
        \multicolumn{2}{c|}{KXNet (ECCV 2022)\cite{kxnet}} & \textcolor{cyan}{\underline{31.96}}& \textcolor{cyan}{\underline{30.67}}& \textcolor{cyan}{\underline{29.57}}& 29.12& 27.84 & \textcolor{cyan}{\underline{26.98}}& 28.24& 27.08& \textcolor{cyan}{\underline{26.26}}& 26.67 & 25.34& 24.49\\
        \multicolumn{2}{c|}{HAT (CVPR 2023)\cite{HAT}} & 31.78& 30.41& 29.30& 29.26& \textcolor{cyan}{\underline{27.93}} & 26.91& 28.25&  \textcolor{cyan}{\underline{27.12}}& 26.22& 26.99 & \textcolor{RubineRed}{\textbf{25.95}}& \textcolor{cyan}{\underline{24.58}}\\
        \multicolumn{2}{c|}{SRFormer (ICCV 2023)\cite{srformer}}  & 31.72& 30.28& 29.17& 29.02& 27.75  & 26.81& 28.15& 27.01& 26.16& 26.46& 25.15& 24.29\\
        \multicolumn{2}{c|}{IPG (CVPR 2024)\cite{IPG}}  & 31.43& 30.19& 28.99& 29.13& 27.76& 26.59& 28.14& 26.97& 26.11& 26.97& 25.54& 24.31\\
        \multicolumn{2}{c|}{CFSR (TIP 2024)\cite{CFSR}}  & 31.94& 30.17& 29.09& 29.14& 27.55 & 26.64& 28.26& 26.90& 26.08& 26.58 & 24.84& 23.97\\
        \multicolumn{2}{c|}{CRAFT (TPAMI 2025)\cite{CRAFT}}  & 31.45& 30.12& 29.07& 29.02& 27.70 & 26.72& 28.14& 26.99& 26.13& 26.67 & 25.19& 24.29 \\
        \hline
        \multicolumn{2}{c|}{\cellcolor{red!5} {\textcolor{RubineRed}{\textbf{KANO (Ours)}}}} & \cellcolor{red!5} {\textcolor{RubineRed}{\textbf{32.29}}}& \cellcolor{red!5} {\textcolor{RubineRed}{\textbf{30.78}}}& \cellcolor{red!5} {\textcolor{RubineRed}{\textbf{29.73}}}& \cellcolor{red!5} {\textcolor{RubineRed}{\textbf{29.36}}}& \cellcolor{red!5} {\textcolor{RubineRed}{\textbf{27.95}}} & \cellcolor{red!5} {\textcolor{RubineRed}{\textbf{27.07}}}& \cellcolor{red!5} {\textcolor{RubineRed}{\textbf{28.38}}}& \cellcolor{red!5} {\textcolor{RubineRed}{\textbf{27.15}}}& \cellcolor{red!5} {\textcolor{RubineRed}{\textbf{26.32}}}& \cellcolor{red!5} {\textcolor{RubineRed}{\textbf{27.16}}} & \cellcolor{red!5} {\textcolor{cyan}{\underline{25.64}}}& \cellcolor{red!5} {\textcolor{RubineRed}{\textbf{24.79}}}\\
        \hline
        \hline
        \multicolumn{2}{c|}{\cellcolor{gray!10} {\textbf{$\mathcal{N}=15$}}} & \cellcolor{gray!10}{\textbf{2}} & \cellcolor{gray!10}{\textbf{3}} &\cellcolor{gray!10}{\textbf{4}} &\cellcolor{gray!10}{\textbf{2}} &\cellcolor{gray!10}{\textbf{3}} & \cellcolor{gray!10}{\textbf{4}} & \cellcolor{gray!10}{\textbf{2}} &\cellcolor{gray!10}{\textbf{3}} &\cellcolor{gray!10}{\textbf{4}} &\cellcolor{gray!10}{\textbf{2}} & \cellcolor{gray!10}{\textbf{3}} & \cellcolor{gray!10}{\textbf{4}}  \\
        \hline
        \hline
        \multicolumn{2}{c|}{EDSR (CVPR 2017)\cite{EDSR}} & 29.68& 28.37& 27.55& 27.36& 26.34 & 25.68& 26.73& 25.86& 25.24& 25.05& 24.00& 23.43\\
        \multicolumn{2}{c|}{IMDN (ACMM 2019)\cite{IMDN}}  & 29.63& 28.35& 27.37& 27.32& 26.30 & 25.53& 26.70& 25.84& 25.17& 24.91& 23.90& 23.12 \\
        \multicolumn{2}{c|}{SwinIR (ICCV 2021)\cite{swinir}}  &  \textcolor{cyan}{\underline{29.94}}&  \textcolor{cyan}{\underline{28.64}}& 27.65& 27.72& 26.49 & 25.72& \textcolor{RubineRed}{\textbf{26.92}}& 25.96& 25.27& \textcolor{cyan}{\underline{25.66}}& 24.31& 23.57\\
        \multicolumn{2}{c|}{ELAN (ECCV 2022)\cite{ELAN}}  &29.86&28.56& \textcolor{cyan}{\underline{27.69}}& \textcolor{cyan}{\underline{27.56}}& 26.50& 25.76& 26.84& 25.96&25.28& 25.33& 24.31& 23.64\\
        \multicolumn{2}{c|}{KXNet (ECCV 2022)\cite{kxnet}} & 29.86& 28.62& 27.64& 27.55& 26.52 & 25.76& 26.83& 25.96&  \textcolor{cyan}{\underline{25.29}}& 25.32 & 24.32& 23.59\\
        \multicolumn{2}{c|}{HAT (CVPR 2023)\cite{HAT}} & 29.92& 28.63& 27.64& 27.68& \textcolor{RubineRed}{\textbf{26.66 }}& \textcolor{cyan}{\underline{25.77}}& 26.88& \textcolor{RubineRed}{\textbf{26.04}}& 25.28& 25.65 &  \textcolor{RubineRed}{\textbf{24.86}}& \textcolor{cyan}{\underline{23.69}}\\
        \multicolumn{2}{c|}{SRFormer (ICCV 2023)\cite{srformer}}  & 29.78& 28.50& 27.52& 27.53& 26.49  & 25.71& 26.81& 25.94& 25.25& 25.25& {{24.24}}& 23.48\\
        \multicolumn{2}{c|}{IPG (CVPR 2024)\cite{IPG}}  & 29.75& 28.52& 27.54& 27.66& 26.58& 25.69& 26.86& 25.98& 25.28& 25.65& 24.60& 23.65\\
        \multicolumn{2}{c|}{CFSR (TIP 2024)\cite{CFSR}}  & 29.86& 28.38& 27.40& 27.55& 26.34 & 25.56& 26.83& 25.86& 25.18& 25.25 & 23.97& 23.21\\
        \multicolumn{2}{c|}{CRAFT (TPAMI 2025)\cite{CFSR}}  & 29.75& 28.44& 27.51& 27.55& 26.48 & 25.67& 26.83& 25.94& 25.26& 25.38 & 24.26& 23.50\\
        \hline
        \multicolumn{2}{c|}{\cellcolor{red!5} {\textcolor{RubineRed}{\textbf{KANO (Ours)}}}} & \cellcolor{red!5} {\textcolor{RubineRed}{\textbf{30.08}}}& \cellcolor{red!5} {\textcolor{RubineRed}{\textbf{28.75}}}& \cellcolor{red!5} {\textcolor{RubineRed}{\textbf{27.79}}}& \cellcolor{red!5} {\textcolor{RubineRed}{\textbf{27.71}}}& \cellcolor{red!5} {\textcolor{cyan}{\underline{26.61}}} & \cellcolor{red!5} {\textcolor{RubineRed}{\textbf{25.86}}}& \cellcolor{red!5} {\textcolor{RubineRed}{\textbf{26.92}}}& \cellcolor{red!5} {\textcolor{cyan}{\underline{26.02}}}& \cellcolor{red!5} {\textcolor{RubineRed}{\textbf{25.33}}}& \cellcolor{red!5} {\textcolor{RubineRed}{\textbf{25.67}}} & \cellcolor{red!5} {\textcolor{cyan}{\underline{24.54}}}& \cellcolor{red!5} {\textcolor{RubineRed}{\textbf{23.80}}}\\
        \Xhline{1.0pt}
    \end{tabular}}
    \label{tab:cv}
\end{table*}
\begin{figure*}[!t]
 	  \centering
 			\includegraphics[width=1.0\textwidth]{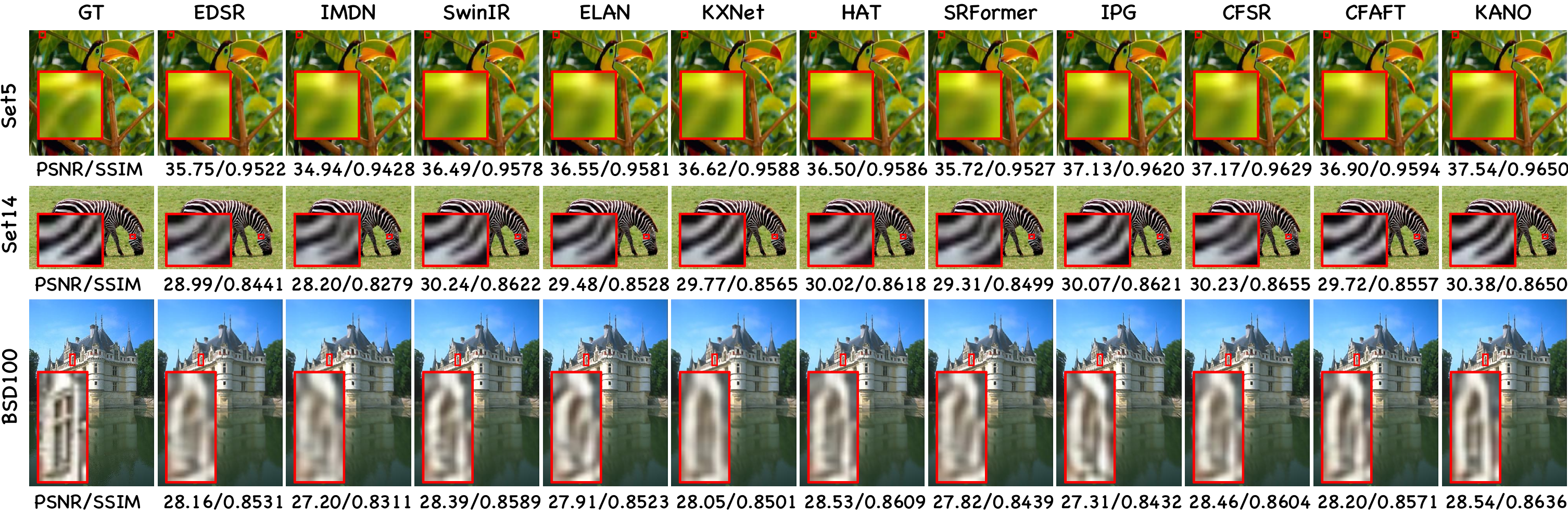}
          \caption{Performance comparison of KANO and other SOTA methods on four benchmark images from Set5 (top), Set14 (upper middle), and BSD100 (bottom), respectively, with the setting of scale factor=2.}
  \label{fig:cv}
  \end{figure*}

The total loss function is utilize the $\mathcal{L}_1$ loss to supervise the predicted degradation kernel $\mathbf{K}^{(t)}$ and the estimated HR image $\mathbf{X}^{(t)}$ at each stage
\begin{equation}
    \begin{aligned}
    \label{eq5.13}
    \mathcal{L}_{total}&=\mathcal{L}_{\mathbf{K}}+\mathcal{L}_{\mathbf{X}}, \\
        &=\int_{\tau}\alpha(\tau)\left\Vert\mathbf{K}-\mathbf{K}^{\tau}\right\Vert_{{\scriptscriptstyle L^1(\Omega,\mu)}}d\nu(\tau)+\sum_{t=1}^{T}\beta_t\left\Vert\mathbf{X}-\mathbf{X}^{(t)}\right\Vert_1, \\ 
        &\Rightarrow \sum_{t=1}^{T}\alpha_t\left\Vert\mathbf{K}-\mathbf{K}^{(t)}\right\Vert_1+\sum_{t=1}^{T}\beta_t\left\Vert\mathbf{X}-\mathbf{X}^{(t)}\right\Vert_1,\\
    \end{aligned}
\end{equation}
where $\left\Vert\mathbf{K}-\mathbf{K}^{\tau}\right\Vert_{\tiny {L^1(\Omega,\mu)}}$ is the norm of the operator $\mathbf{K}$ based on the measure $\mu$ in space $\Omega$, and convert to the norm for the degenerate kernel for simplicity at third line.

\begin{table*}[!t]
    \centering
\caption{A quantitative comparison of average PSNR (dB) among different SR methods, evaluated on a widely \textit{\textcolor{RubineRed}{aerial image}} (CAVE), with five blur factors $\sigma \in \{1.5, 1.8, 1.95, 2.1, 2.4 \}$ and three sampling scale $\gamma \in \{2, 4, 8\}$. Throughout this paper, the best and second-best results of each test case are highlighted in \textcolor{RubineRed}{\textbf{Bold}}, \textcolor{cyan}{\underline{blue}}, respectively.}
\scalebox{0.8}{
    \begin{tabular}{c|c|ccccc|ccccc|ccccc}
        \Xhline{1.0pt}
         \cellcolor{gray!10}{}&\cellcolor{gray!10}{\textbf{$\gamma$}} & \multicolumn{5}{c|}{\cellcolor{gray!10}{\textbf{$\times$ 2 }}} & \multicolumn{5}{c|}{\cellcolor{gray!10}{\textbf{$\times$ 4 }}} & \multicolumn{5}{c}{\cellcolor{gray!10}{\textbf{$\times$ 8 }}} \\
          \hhline{>{\arrayrulecolor{gray!10}}->{\arrayrulecolor{black}}|-|-|-|-|-|-|-|-|-|-|-|-|-|-|-|-|}
             \multirow{-2}{*}{\textbf{\cellcolor{gray!10}{\textbf{Method}}}} & \multicolumn{1}{c|}{\cellcolor{gray!10} {\textbf{$\sigma$}}} & \cellcolor{gray!10}{\textbf{1.5}} & \cellcolor{gray!10}{\textbf{1.8}} &\cellcolor{gray!10}{\textbf{1.95}} &\cellcolor{gray!10}{\textbf{2.1}} &\cellcolor{gray!10}{\textbf{2.4}} & \cellcolor{gray!10}{\textbf{1.5}} & \cellcolor{gray!10}{\textbf{1.8}} &\cellcolor{gray!10}{\textbf{1.95}} &\cellcolor{gray!10}{\textbf{2.1}} &\cellcolor{gray!10}{\textbf{2.4}} & \cellcolor{gray!10}{\textbf{1.5}} & \cellcolor{gray!10}{\textbf{1.8}} &\cellcolor{gray!10}{\textbf{1.95}} &\cellcolor{gray!10}{\textbf{2.1}} &\cellcolor{gray!10}{\textbf{2.4}}  \\
        \hline
        \hline
        \multicolumn{17}{c}{\cellcolor{gray!15}{\textbf{CAVE ( 32 aerial images of size 512 $\times$ 512 with 31 Channels )}}}\\
        \hline
        \hline
       \multicolumn{2}{c|}{Biubic} &35.54 &35.16 &34.80 &34.44 &34.11 &31.51 &31.50 &31.47 &31.42 &31.37 &27.16 &27.23 &27.29 &27.34 &27.39\\
        \multicolumn{2}{c|}{GDRRN\cite{Li2018GDRRN}}  &37.90 &37.34 &36.74 &36.15 &35.58 &35.04 &34.95 &34.80 &34.60 &34.36 &30.20 &30.26 &30.31 &30.35 &30.36 \\
        \multicolumn{2}{c|}{SSPSR\cite{Jiang2020SSPSR}}  &40.19 &40.09 &39.75 &39.05 &38.14 &36.96 &36.83 &36.61 &36.31 &35.95 &32.15 &32.26 &32.34 &32.37 &32.37 \\
        \multicolumn{2}{c|}{ERCSR\cite{9334383}}  & 40.80& \textcolor{cyan}{\underline{40.83}} & \textcolor{cyan}{\underline{40.74}}& 40.19& 39.12& 36.87& 36.67& 36.40&36.08& 35.72& 32.13& 32.24& 32.32& 32.36 &32.37\\
        \multicolumn{2}{c|}{3DQRNN\cite{Fu20213DQRNN}} &39.87 &39.76 &39.37 &38.65 &37.74 &36.42 &36.28 &36.07 &35.79 &35.47 &31.46 &31.56 &31.62 &31.66 &31.67\\
        \multicolumn{2}{c|}{RFSR\cite{wang2021RFSR}}& 38.92& 38.25& 37.57& 36.91& 36.28& 35.96& 35.73& 35.46& 35.16& 34.84& 31.26& 31.34& 31.41& 31.45& 31.47\\
        \multicolumn{2}{c|}{PDENet\cite{hou22pde}}  & 39.95& 39.48& 39.06& 38.36& 37.59& \textcolor{cyan}{\underline{37.17}}& \textcolor{cyan}{\underline{37.05}}& 36.80& 36.51& 36.18& 32.37& \textcolor{cyan}{\underline{32.49}}& \textcolor{cyan}{\underline{32.55}}& \textcolor{cyan}{\underline{32.59}}& \textcolor{cyan}{\underline{32.50}}\\
        \multicolumn{2}{c|}{GELIN\cite{wang2022group}}  & 39.90& 39.54& 39.03& 38.31& 37.45& 37.09& 36.99& 36.81& 35.54& 36.18& 32.21& 32.34& 32.43& 32.47& 32.48\\
        \multicolumn{2}{c|}{ESSAFormer\cite{Zhang2023ESSA}}  &\textcolor{cyan}{\underline{40.86}} &40.81&\textcolor{cyan}{\underline{40.74}} &\textcolor{cyan}{\underline{40.46}} &\textcolor{cyan}{\underline{39.63}} &36.96 &37.01 &\textcolor{cyan}{\underline{36.99}} &\textcolor{cyan}{\underline{36.88}} &\textcolor{cyan}{\underline{36.64}} &\textcolor{cyan}{\underline{32.39}} &32.43 &32.44 &32.41 &32.36\\
        \multicolumn{2}{c|}{SNLSR\cite{hu2024SNLSR}}  &40.51 &40.31 &40.14 &39.86 &39.17 &36.48 &36.50 &36.49 &36.40 &36.22 &31.84 &31.91 &31.93 &31.91 &31.88 \\
        \hline
        \multicolumn{2}{c|}{\cellcolor{red!5} {\textcolor{RubineRed}{\textbf{KANO (Ours)}}}} & \cellcolor{red!5} {\textcolor{RubineRed}{\textbf{41.32}}}& \cellcolor{red!5} {\textcolor{RubineRed}{\textbf{41.08}}}& \cellcolor{red!5} {\textcolor{RubineRed}{\textbf{40.88}}}& \cellcolor{red!5} {\textcolor{RubineRed}{\textbf{40.59}}}& \cellcolor{red!5} {\textcolor{RubineRed}{\textbf{39.91}}} & \cellcolor{red!5} {\textcolor{RubineRed}{\textbf{37.20}}}& \cellcolor{red!5} {\textcolor{RubineRed}{\textbf{37.24}}}& \cellcolor{red!5} {\textcolor{RubineRed}{\textbf{37.23}}}& \cellcolor{red!5} {\textcolor{RubineRed}{\textbf{37.16}}}& \cellcolor{red!5} {\textcolor{RubineRed}{\textbf{36.98}}} & \cellcolor{red!5} {\textcolor{RubineRed}{\textbf{32.53}}}& \cellcolor{red!5} {\textcolor{RubineRed}{\textbf{32.60}}}& \cellcolor{red!5} {\textcolor{RubineRed}{\textbf{32.64}}}& \cellcolor{red!5} {\textcolor{RubineRed}{\textbf{32.63}}}& \cellcolor{red!5} {\textcolor{RubineRed}{\textbf{32.58}}}\\
        \Xhline{1.0pt}
    \end{tabular}}
    \label{tab:cave}
\end{table*}
\begin{figure*}[!t]
 	  \centering
 			\includegraphics[width=1.0\textwidth]{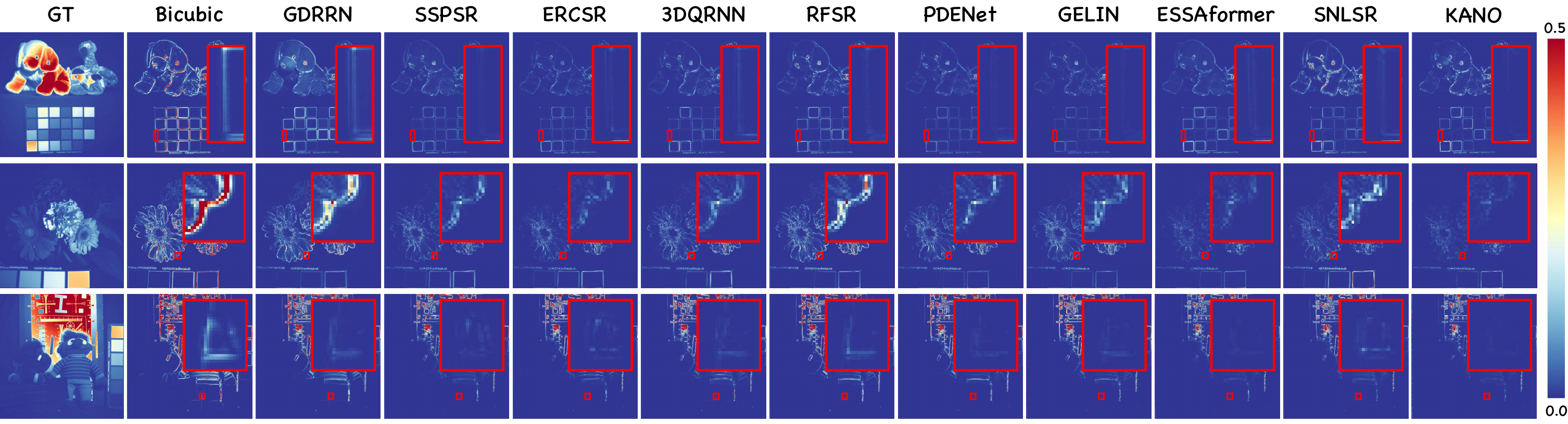}
          \caption{Performance comparison of KANO and other SOTA methods on the Mean Squared Error (MSE) of 3 different scenes ($\sigma$=1.5) in the CAVE dataset.}
  \label{fig:cave}
  \end{figure*}

\section{Experiments}\label{sec:Experiments}
In this section, we provide a comprehensive evaluation on computer vision images and remote sensing images. Performance is assessed using two quantitative metrics, including the Peak Signal-to-Noise Ratio (PSNR) and Structural Similarity Index Measure (SSIM), where higher values indicate better results. All experiments are conducted using the PyTorch framework, with model training performed on eight NVIDIA L40 GPUs.

In addition to benchmarking against state-of-the-art algorithms, we conducted extensive ablation studies to comprehensively validate the proposed innovations, including the KAN-based blur kernel and image spectral modeling, as well as the nonlinear correction of image information. Finally, we compared the complexity and performance of KAN with that of a less controllable MLP within our framework, further substantiating the potential and promise of KAN for SISR tasks.

\subsection{Application in Natural Image}
\subsubsection{Experimental Datasets} 
For natural image super-resolution, we evaluate our model on four widely used benchmark datasets: Set5~\cite{set5}, Set14~\cite{set14}, BSD100~\cite{bsd100}, and Urban100~\cite{urban100}. The model is trained using a combination of the DIV2K and Flickr2K~\cite{Div2k} datasets.

To thoroughly assess the performance of the proposed model under various degradation conditions, we introduce a range of synthetic degradations to the aforementioned datasets. Specifically, three downsampling scales are considered: $\times$2, $\times$3, and $\times$4. The corresponding blur kernel sizes are set to $11 \times 11$, $15 \times 15$, and $21 \times 21$, respectively. For all scales, the kernel variance is randomly sampled from the range [0.6, 5.0]. The kernels are anisotropic, with orientation angles uniformly sampled from $[-\pi, \pi]$. In addition, additive Gaussian noise with a maximum noise level of $25/255$ is introduced to simulate realistic degradation.

During training, input images are randomly cropped into $64 \times 64$ patches. The network is optimized using the Adam optimizer with parameters $\beta_1 = 0.9$ and $\beta_2 = 0.999$, targeting the total loss $\mathcal{L}_{\text{total}}$. The total number of training iterations is 800K, with a batch size of 8. For evaluation, the model is tested on the three degradation scales mentioned above. For each scale, blur kernels are rotated by angles of $0$, $\pi/4$, $\pi/2$, and $3\pi/4$, and Gaussian noise with noise levels of 0/255, 5/255, and 15/255 is added. To ensure fair and reproducible comparisons, all test data are generated and kept fixed throughout the evaluation process.

\begin{table*}[!t]
    \centering
\caption{A quantitative comparison of average PSNR (dB) among different SR methods, evaluated on two widely \textit{\textcolor{RubineRed}{satellite images}} (Pavia and Chikusei), with five blur factors $\sigma \in \{1.5, 1.8, 1.95, 2.1, 2.4 \}$ and three sampling scale $\gamma \in \{2, 4, 8\}$. Throughout this paper, the best and second-best results of each test case are highlighted in \textcolor{RubineRed}{\textbf{Bold}}, \textcolor{cyan}{\underline{blue}}, respectively.}
\scalebox{0.83}{
    \begin{tabular}{c|c|ccccc|ccccc|ccccc}
        \Xhline{1.0pt}
         \cellcolor{gray!10}{}&\cellcolor{gray!10}{\textbf{$\gamma$}} & \multicolumn{5}{c|}{\cellcolor{gray!10}{\textbf{$\times$ 2}}} & \multicolumn{5}{c|}{\cellcolor{gray!10}{\textbf{$\times$ 4}}} & \multicolumn{5}{c}{\cellcolor{gray!10}{\textbf{$\times$ 8 }}} \\
          \hhline{>{\arrayrulecolor{gray!10}}->{\arrayrulecolor{black}}|-|-|-|-|-|-|-|-|-|-|-|-|-|-|-|-|}
             \multirow{-2}{*}{\textbf{\cellcolor{gray!10}{\textbf{Method}}}} & \multicolumn{1}{c|}{\cellcolor{gray!10} {\textbf{$\sigma$}}} & \cellcolor{gray!10}{\textbf{1.5}} & \cellcolor{gray!10}{\textbf{1.8}} &\cellcolor{gray!10}{\textbf{1.95}} &\cellcolor{gray!10}{\textbf{2.1}} &\cellcolor{gray!10}{\textbf{2.4}} & \cellcolor{gray!10}{\textbf{1.5}} & \cellcolor{gray!10}{\textbf{1.8}} &\cellcolor{gray!10}{\textbf{1.95}} &\cellcolor{gray!10}{\textbf{2.1}} &\cellcolor{gray!10}{\textbf{2.4}} & \cellcolor{gray!10}{\textbf{1.5}} & \cellcolor{gray!10}{\textbf{1.8}} &\cellcolor{gray!10}{\textbf{1.95}} &\cellcolor{gray!10}{\textbf{2.1}} &\cellcolor{gray!10}{\textbf{2.4}}  \\
        \hline
        \hline
        \multicolumn{17}{c}{\cellcolor{gray!10}{\textbf{Pavia ( 12 Urban RS images of size 224 $\times$ 224 with 102 Channels )}}}\\
        \hline
        \hline
         \multicolumn{2}{c|}{Biubic} & 28.43& 28.12&28.81& 27.54& 27.28 & 25.81&28.81& 25.80& 25.77& 25.74 & 23.25&23.33 &23.40& 23.47& 23.53\\
        \multicolumn{2}{c|}{GDRRN\cite{Li2018GDRRN}}  & 32.10&31.85& 31.43& 30.80& 30.05 & 28.33&28.23& 28.09& 27.92& 27.73& 24.69&24.75& 24.80& 24.83& 24.84 \\
        \multicolumn{2}{c|}{SSPSR\cite{Jiang2020SSPSR}}  & 33.62&33.41& 33.15& 32.70& 31.86 & 28.92&28.94 &28.93& 28.87& 28.75& 25.19&25.22 &25.24& 25.24& 25.23 \\
        \multicolumn{2}{c|}{ERCSR\cite{9334383}}  & {{34.89}}&34.81& 34.69& 34.36& 33.26& 28.92&\textcolor{cyan}{\underline{28.96}} &\textcolor{cyan}{\underline{28.97}}& \textcolor{cyan}{\underline{28.93}}& 28.84& \textcolor{cyan}{\underline{25.29}}&\textcolor{cyan}{\underline{25.34}} &\textcolor{cyan}{\underline{25.36}}& 25.37& 25.37\\
        \multicolumn{2}{c|}{3DQRNN\cite{Fu20213DQRNN}} & 30.20&29.60 &29.06& 28.60& 28.18 & 28.27&28.16 &28.01& 27.83& 27.64 & 25.10&25.14 &25.16& 25.17& 25.16\\
        \multicolumn{2}{c|}{RFSR\cite{wang2021RFSR}} & 32.94&32.69 &32.41& 31.82& 30.83 & 28.80&28.83 &28.81& 28.72& 28.56 & 25.11&25.16 &25.19& 25.21& 25.21\\
        \multicolumn{2}{c|}{PDENet\cite{hou22pde}}  & \textcolor{cyan}{\underline{35.25}}&\textcolor{cyan}{\underline{35.17}} &\textcolor{cyan}{\underline{34.98}}& \textcolor{cyan}{\underline{34.68}}& \textcolor{cyan}{\underline{33.75}} & 28.77&28.81 &28.84& 28.82& 28.82& 25.26&25.33 &\textcolor{cyan}{\underline{25.36}}& \textcolor{cyan}{\underline{25.38}}& \textcolor{cyan}{\underline{25.38}}\\
        \multicolumn{2}{c|}{GELIN\cite{wang2022group}}  & 34.43&34.36 &34.20& 33.76& 32.68 & 28.80&28.86 &28.91 & 28.92& \textcolor{cyan}{\underline{28.90}}& 25.08&25.15 &25.20& 25.25& 25.28\\
        \multicolumn{2}{c|}{ESSAformer\cite{Zhang2023ESSA}}  &34.39&34.26&34.06&33.76&33.29 & \textcolor{cyan}{\underline{28.93}}&28.94 &28.91& 28.83& 28.69 & 25.16&25.21 &25.23& 25.24& 25.24\\
        \multicolumn{2}{c|}{SNLSR\cite{hu2024SNLSR}}  & 34.62&34.46 &34.25& 33.91& 32.94 & 28.83&28.86 &28.86& 28.86& 28.80 & 25.16&25.20 &25.23& 25.23& 25.24\\
        \hline
        \multicolumn{2}{c|}{\cellcolor{red!5} {\textcolor{RubineRed}{\textbf{KANO (Ours)}}}} & \cellcolor{red!5} {\textcolor{RubineRed}{\textbf{35.44}}}&\cellcolor{red!5} {\textcolor{RubineRed}{\textbf{35.30}}} &\cellcolor{red!5} {\textcolor{RubineRed}{\textbf{35.10}}}& \cellcolor{red!5} {\textcolor{RubineRed}{\textbf{34.84}}}& \cellcolor{red!5} {\textcolor{RubineRed}{\textbf{34.31}}} & \cellcolor{red!5} {\textcolor{RubineRed}{\textbf{29.17}}}&\cellcolor{red!5} {\textcolor{RubineRed}{\textbf{29.18}}} &\cellcolor{red!5} {\textcolor{RubineRed}{\textbf{29.14}}}& \cellcolor{red!5} {\textcolor{RubineRed}{\textbf{29.07}}}& \cellcolor{red!5} {\textcolor{RubineRed}{\textbf{28.93}}}& \cellcolor{red!5} {\textcolor{RubineRed}{\textbf{25.35}}}&\cellcolor{red!5} {\textcolor{RubineRed}{\textbf{25.40}}} &\cellcolor{red!5} {\textcolor{RubineRed}{\textbf{25.44}}}& \cellcolor{red!5} {\textcolor{RubineRed}{\textbf{25.46}}}& \cellcolor{red!5} {\textcolor{RubineRed}{\textbf{25.46}}}\\
        \hline
        \hline
        \multicolumn{17}{c}{\cellcolor{gray!10}{  \textbf{Chikusei ( 16 satellite images of size 256 $\times$ 256 with 128 Channels )}}}\\
        \hline
        \hline
        \multicolumn{2}{c|}{Biubic} & 38.05&37.68 &37.32& 36.99& 36.69 & 34.99&35.01 &34.99& 34.97& 34.93& 32.27&32.36 &32.44& 32.51& 32.57 \\
        \multicolumn{2}{c|}{GDRRN\cite{Li2018GDRRN}}  & 43.22&43.04 &42.66& 41.91& 40.88 & 37.95&37.87 &37.74& 37.58& 37.38& 34.19&34.26 &34.30& 34.33& 34.35 \\
        \multicolumn{2}{c|}{SSPSR\cite{Jiang2020SSPSR}}  & 44.53&44.28 &43.96& 43.38& 42.16 & 38.84&38.89 &38.87& 38.80& 38.65& \textcolor{cyan}{\underline{34.42}}&\textcolor{cyan}{\underline{34.49}} &\textcolor{cyan}{\underline{34.52}}& \textcolor{cyan}{\underline{34.55}}& \textcolor{cyan}{\underline{34.56}} \\
        \multicolumn{2}{c|}{ERCSR\cite{9334383}}  & 45.89&45.80 &45.66& 45.25& 43.83& 38.76&38.82 &38.83& 38.79& 38.66& 34.36&34.41 &34.45& 34.48& 34.49 \\
        \multicolumn{2}{c|}{3DQRNN\cite{Fu20213DQRNN}} & 39.83&39.13 &38.51& 37.97& 37.50 & 37.78&37.67 &37.50& 37.30& 37.08& 34.16&34.21 &34.23& 34.24& 34.24 \\
        \multicolumn{2}{c|}{RFSR\cite{wang2021RFSR}} & 45.37&45.16 &44.85& 44.25& 42.73 & 38.68&38.73 &38.72& 38.62& 38.41 & 34.41&34.46& 34.50& 34.52& 34.52 \\
        \multicolumn{2}{c|}{PDENet\cite{hou22pde}} & \textcolor{cyan}{\underline{46.28}}&46.23& 45.85& 45.02& 43.01 &38.92 &38.99& 39.04& 39.06& 38.98&34.24 &34.33& 34.37& 34.41& 34.45 \\
        \multicolumn{2}{c|}{GELIN\cite{wang2022group}}  & 46.61&\textcolor{cyan}{\underline{46.59}} &\textcolor{cyan}{\underline{46.45}}& \textcolor{cyan}{\underline{46.20}}& \textcolor{cyan}{\underline{44.69}}& 38.98&39.05 &39.07& 39.04& 38.89& 34.31&34.39 &34.46& 34.51& \textcolor{cyan}{\underline{34.56}} \\
        \multicolumn{2}{c|}{ESSAformer\cite{Zhang2023ESSA}} & 43.01&42.36 &41.57& 40.69& 39.79 & 38.73&38.75 &38.70& 38.57& 38.34& 34.32&34.37 &34.41& 34.43& 34.44 \\
        \multicolumn{2}{c|}{SNLSR\cite{hu2024SNLSR}}  & 45.96&45.76 &45.49& 45.06& 43.85& \textcolor{cyan}{\underline{39.04}}&\textcolor{cyan}{\underline{39.11}}& \textcolor{cyan}{\underline{39.14}}& \textcolor{cyan}{\underline{39.12}}& \textcolor{cyan}{\underline{38.99}} & 34.19&34.28 &34.36& 34.43& 34.48 \\
        \hline
        \multicolumn{2}{c|}{\cellcolor{red!5} {\textcolor{RubineRed}{\textbf{KANO (Ours)}}}} & \cellcolor{red!5} {\textcolor{RubineRed}{\textbf{47.30}}}&\cellcolor{red!5} {\textcolor{RubineRed}{\textbf{47.14}}}& \cellcolor{red!5} {\textcolor{RubineRed}{\textbf{46.92}}}& \cellcolor{red!5} {\textcolor{RubineRed}{\textbf{46.57}}}& \cellcolor{red!5} {\textcolor{RubineRed}{\textbf{45.77}}}& \cellcolor{red!5} {\textcolor{RubineRed}{\textbf{39.37}}}&\cellcolor{red!5} {\textcolor{RubineRed}{\textbf{39.43}}}& \cellcolor{red!5} {\textcolor{RubineRed}{\textbf{39.45}}}& \cellcolor{red!5}{\textcolor{RubineRed}{\textbf{39.42}}}& \cellcolor{red!5} {\textcolor{RubineRed}{\textbf{39.30}}} & \cellcolor{red!5} {\textcolor{RubineRed}{\textbf{34.51}}}&\cellcolor{red!5} {\textcolor{RubineRed}{\textbf{34.58}}} &\cellcolor{red!5} {\textcolor{RubineRed}{\textbf{34.63}}}& \cellcolor{red!5} {\textcolor{RubineRed}{\textbf{34.67}}}& \cellcolor{red!5} {\textcolor{RubineRed}{\textbf{34.69}}}\\
        \Xhline{1.0pt}
    \end{tabular}}
    \label{tab:Pavia_Chikusei}
\end{table*}
\begin{figure*}[!t]
 	  \centering
 			\includegraphics[width=1.0\textwidth]{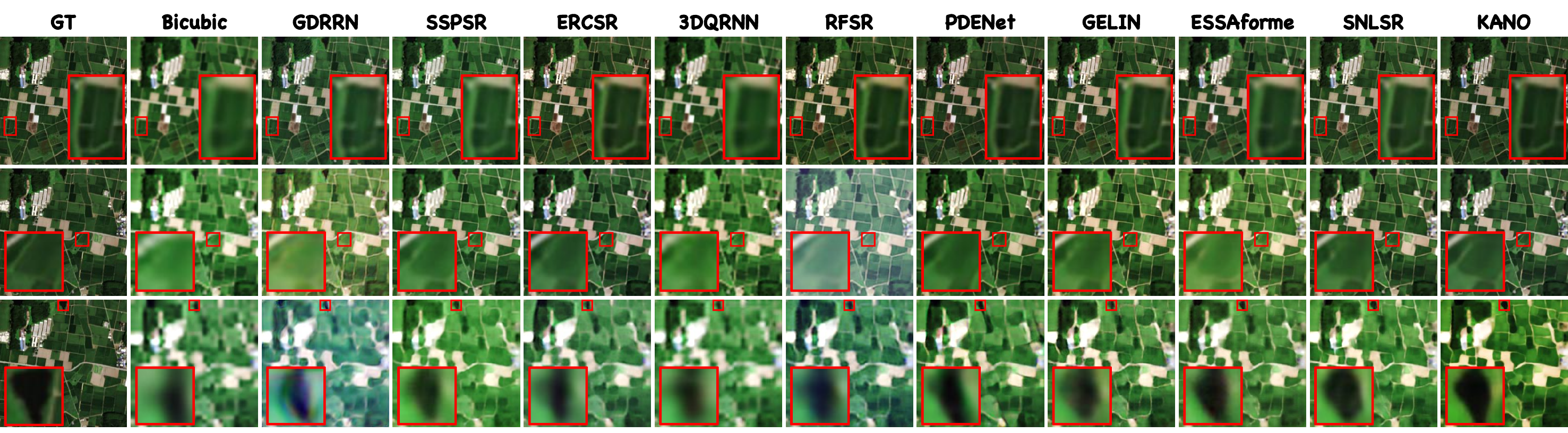}
          \caption{Performance comparison of KANO and other SOTA methods on the false-color image (50, 40, 30) of 3 same scenes (scale factor=2,4,8) in the Chikusei dataset.  }
  \label{fig:Chikusei}
  \end{figure*}

\subsubsection{Experimental Comparison.} 
Tab. \ref{tab:cv} presents a detailed comparison of the image reconstruction performance of various algorithms across four datasets under scale degradations of ×2, ×3, and ×4, further compounded by three levels of noise severity. As shown in the results, KANO consistently achieves the best performance on the Set5, Set14, and BSD100 datasets under noise levels 0 and 5. Although it is slightly outperformed by HAT on the Urban100 dataset under ×3 degradation, KANO demonstrates superior performance at both smaller (×2) and larger (×4) scales, highlighting its robustness on more complex and diverse scenes. Moreover, as the noise level increases, all super-resolution methods suffer from varying degrees of performance degradation. Nevertheless, KANO maintains superior results across all datasets, even under extremely high noise levels, demonstrating strong generalization capability and noise robustness.

Fig. \ref{fig:cv} provides a visual comparison of the super-resolved outputs generated by different methods. On fine-grained facial regions, such as eyelashes, both KXNet and KANO exhibit clear foreground-background separation and recover richer texture details. A similar advantage is observed in the BSD100 dataset, where these two methods again provide more accurate structural reconstruction. In contrast, the Set14 results reveal evident checkerboard artifacts in CRAFT, while CFSR fails to recover the prominent diagonal black line from the top-left to the bottom-right. KANO and IPG, however, produce the most perceptually faithful results. On the Urban100 dataset, many competing methods show noticeable diagonal spatial distortions, whereas KANO successfully preserves geometric consistency and reconstructs high-quality images.


\begin{figure*}[!t]
 	  \centering
			\includegraphics[width=1.0\textwidth]{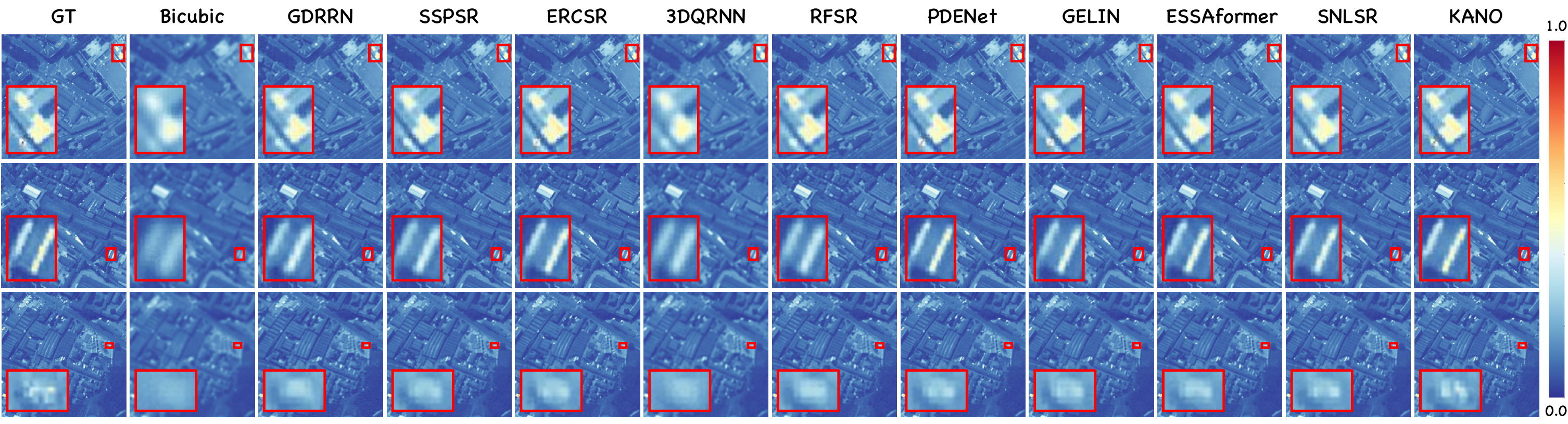}
          \caption{Performance comparison of KANO and other SOTA methods on the ninth band of 3 different scenes (scale factor=2) in the Pavia dataset.}
  \label{fig:pavia}
  \end{figure*}
\begin{figure*}[!t]
 	  \centering
 			\includegraphics[width=1.0\textwidth]{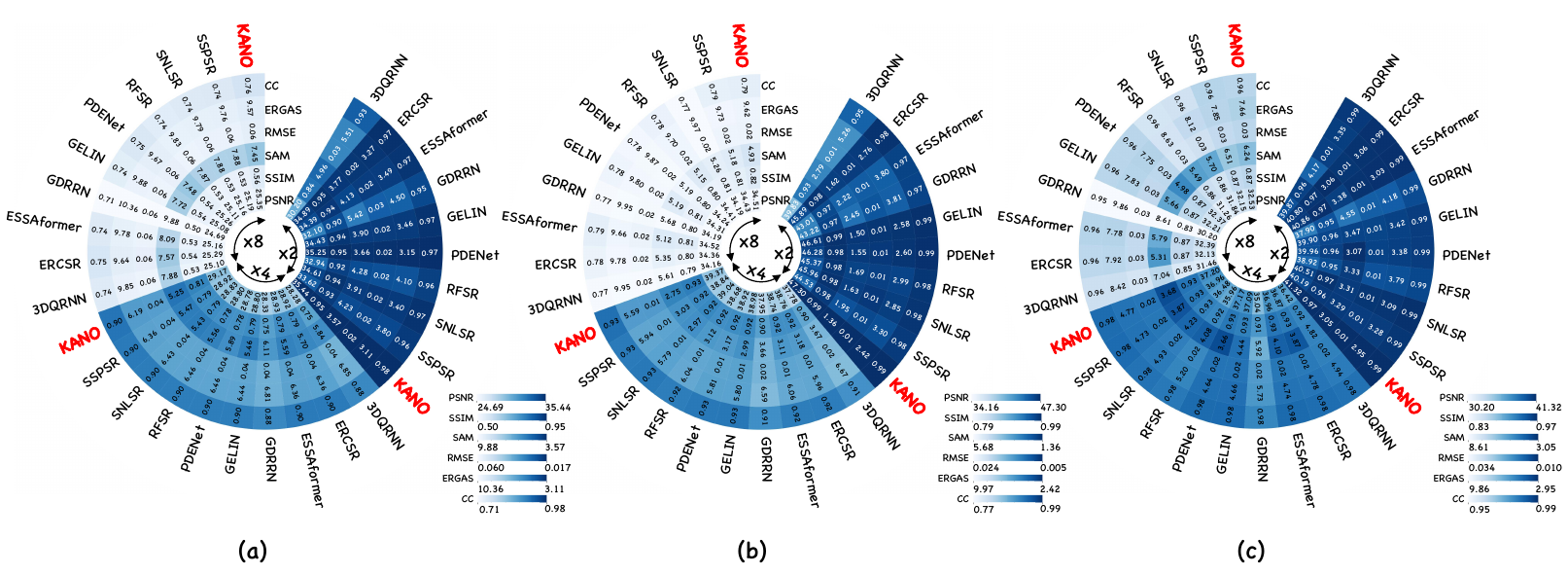}
          \caption{Quantitative comparison between KAN and competing methods on the Pavia, Chikusei, and CAVE datasets under three degradation scales ($\times$2, $\times$4, and $\times$8). Subfigures (a), (b), and (c) correspond to results on Pavia, Chikusei, and CAVE, respectively. The metrics include PSNR, SSIM, SAM, RMSE, ERGAS, and Cross-Correlation (CC).}
  \label{fig:abl_cyc}
  \end{figure*}

\subsection{Application in Remote Sensing}
\subsubsection{Experimental Datasets} 
Hyperspectral images (HSIs), comprising hundreds of contiguous spectral bands, exhibit significant spatial heterogeneity and strong spectral correlations, which further increase the complexity of hyperspectral super-resolution (HSR). Our experiments are conducted on the following three widely used datasets:

\textbf{CAVE}: This dataset was captured using a tunable filter and a cooled CCD camera over the spectral range of 400-700 nm at 10 nm intervals. It contains 31 indoor images of size 512$\times$512$\times$31, covering a diverse set of richly textured scenes. In this study, 21 images are used for training and the remaining 10 for testing.

\textbf{Pavia}: Collected at the University of Pavia, Italy, this benchmark hyperspectral dataset consists of 1096$\times$715 pixels across 102 spectral bands, with a spatial resolution of approximately 1.3 meters per pixel. Each image is partitioned into 12 non-overlapping blocks of size 224$\times$224$\times$102, of which nine blocks are used for training and the remaining three for testing.

\textbf{Chikusei}: Acquired in Chikusei, Ibaraki, Japan, on 29 July 2014, this airborne hyperspectral data set covers a spectral range of 360-1018 nm and has a spatial size of 2517$\times$2335$\times$128. The dataset is divided into 16 patches of size 256×256 pixels; 12 patches are used for training, and the remaining 4 are reserved for testing.

To ensure a fair comparison, all methods are trained and evaluated using the same training and testing splits, along with identical data pre-processing procedures. Furthermore, each model is implemented with the official parameter settings provided in their respective publications.

\begin{figure*}[!t]
 	  \centering
 			\includegraphics[width=1.0\textwidth]{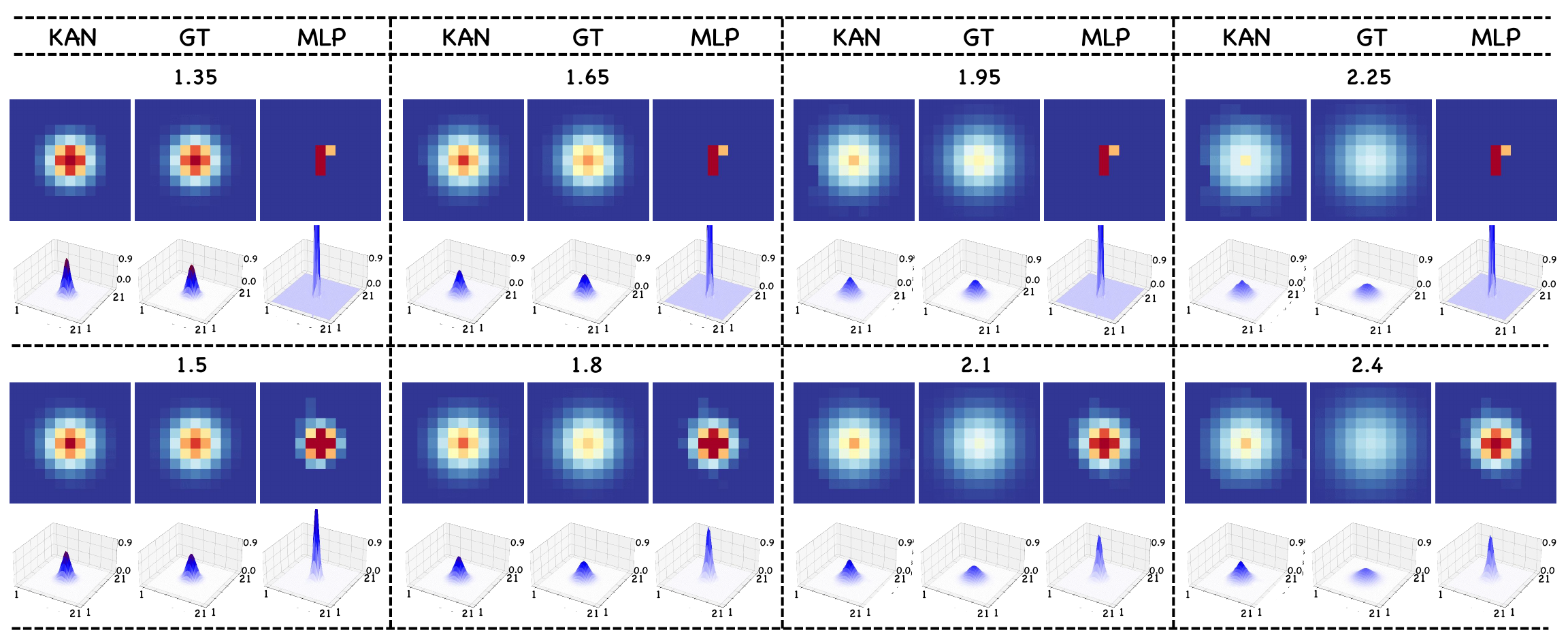}
          \caption{lur kernel estimation results of KANO using KAN and MLP backbones. The top row shows results on the Pavia dataset with kernel variances of 1.35, 1.65, 1.95, and 2.25; the bottom row shows results on the CAVE dataset with variances of 1.5, 1.8, 2.1, and 2.4. KANO with KAN produces kernels closer to the ground truth, while MLP-based results exhibit larger outliers, demonstrating the advantage of KAN in kernel modeling.}
  \label{fig:mlp_kernel}
  \end{figure*}
\begin{table*}[!t]
    \centering
\caption{Ablation analysis of the proposed KNAO foundation model in various layers of KAN in terms of DIV2K, Set5, Urban100, CAVE, Pavia, and Chikusei, respectively. Throughout this paper, the best and second-best results of each test case are highlighted in \textcolor{RubineRed}{\textbf{Bold}}, \textcolor{cyan}{\underline{blue}}, respectively.}
\scalebox{0.95}{
    \begin{tabular}{c|cc|cc|cc|cc|cc|cc}
	\Xhline{1.0pt}
         \cellcolor{gray!10}{\textbf{}} & \multicolumn{2}{c|}{\cellcolor{gray!10} {\textbf{DIV2K}}}& \multicolumn{2}{c|}{\cellcolor{gray!10} {\textbf{Set5}}} & \multicolumn{2}{c|}{\cellcolor{gray!10} {\textbf{Urban100}}}& \multicolumn{2}{c|}{\cellcolor{gray!10} {\textbf{CAVE}}}& \multicolumn{2}{c|}{\cellcolor{gray!10} {\textbf{Pavia}}} & \multicolumn{2}{c}{\cellcolor{gray!10} {\textbf{Chikusei}}}\\
          \hhline{>{\arrayrulecolor{gray!10}}->{\arrayrulecolor{black}}|-|-|-|-|-|-|-|-|-|-|-|-}
             \multirow{-2}{*}{{\cellcolor{gray!10}{\textbf{Depth}}}} & \cellcolor{gray!10}{\textbf{PSNR}} & \cellcolor{gray!10}{\textbf{SSIM}} & \cellcolor{gray!10}{\textbf{PSNR}} & \cellcolor{gray!10}{\textbf{SSIM}} & \cellcolor{gray!10}{\textbf{PSNR}} & \cellcolor{gray!10}{\textbf{SSIM}} & \cellcolor{gray!10}{\textbf{PSNR}} & \cellcolor{gray!10}{\textbf{SSIM}} & \cellcolor{gray!10}{\textbf{PSNR}} & \cellcolor{gray!10}{\textbf{SSIM}}& \cellcolor{gray!10}{\textbf{PSNR}} & \cellcolor{gray!10}{\textbf{SSIM}}\\
        \hline
        \hline
        \multicolumn{1}{c|}{1} &\textcolor{cyan}{\underline{28.80}} &\textcolor{cyan}{\underline{0.8014}} &\textcolor{RubineRed}{\textbf{34.88}} &\textcolor{cyan}{\underline{0.9314}} &\textcolor{cyan}{\underline{28.45}} &\textcolor{cyan}{\underline{0.8625}} &\textcolor{RubineRed}{\textbf{40.60}} &\textcolor{RubineRed}{\textbf{0.9647}} &34.86 &0.9455 &\textcolor{RubineRed}{\textbf{46.88}} &\textcolor{RubineRed}{\textbf{0.9870}}\\
        \multicolumn{1}{c|}{2}  &28.74 &0.8013 &34.62 &0.9297 &28.39 &0.8611 &\textcolor{cyan}{\underline{40.57}} &\textcolor{cyan}{\underline{0.9631}} &\textcolor{RubineRed}{\textbf{35.01}} &\textcolor{RubineRed}{\textbf{0.9470}} & \textcolor{cyan}{\underline{46.87}} &\textcolor{RubineRed}{\textbf{0.9870}}\\
        \multicolumn{1}{c|}{3}  &\textcolor{RubineRed}{\textbf{28.82}} &\textcolor{RubineRed}{\textbf{0.8021}} &\textcolor{cyan}{\underline{34.84}} &\textcolor{RubineRed}{\textbf{0.9318}} &\textcolor{RubineRed}{\textbf{28.49}} &\textcolor{RubineRed}{\textbf{0.8636}} &39.56 &0.9587 &\textcolor{cyan}{\underline{34.98}} &0.9461 &46.40 &0.9860\\
        \multicolumn{1}{c|}{4}  & 28.73& 0.7998& 34.68 &0.9296 &28.30 & 0.8584 &40.54& 0.9615 &34.97 & \textcolor{cyan}{\underline{0.9469}} &46.70 &0.9868\\
	\Xhline{1.0pt}
	\end{tabular}}
    \label{tab:abl_depth}
\end{table*}

\subsubsection{Experimental Comparison} 
Hyperspectral images exhibit strong spatial heterogeneity and complex spectral characteristics, making hyperspectral super-resolution (HSR) particularly challenging. Tab. \ref{tab:cave}, Tab. \ref{tab:Pavia_Chikusei}, and Fig. \ref{fig:abl_cyc} report the performance of various algorithms on three datasets under different blur kernels, with scale degradation factors of ×2, ×4, and ×8, respectively. On the CAVE dataset, ESSA achieves the best results among the compared methods under ×2 and ×4 degradation, while PDENet performs best under ×8 degradation. In comparison, KANO consistently outperforms both methods across all degradation levels, achieving PSNR gains ranging from 0.04 dB to 0.46 dB and delivering the best overall performance. 

Tab. \ref{tab:Pavia_Chikusei} also presents results on the Pavia and Chikusei datasets. On Pavia, KANO achieves the best performance across all degradation scales. Its strong results on this urban-focused dataset demonstrate its improved ability to capture spatial detail and handle scenarios with high complexity and varied object scales. For the Chikusei dataset, while GELIN, SNLSR, and SSPSR show competitive performance at degradation factors of 2, 4, and 8 respectively, KANO achieves PSNR improvements ranging from 0.09 dB to 1.08 dB, again obtaining the best overall results. These findings confirm KANO’s strong fitting ability and effectiveness in scenarios with high spectral similarity and relatively low spatial resolution.


To further support these results, we visualize three representative reconstruction examples in Figs. \ref{fig:cave}, \ref{fig:pavia}, and \ref{fig:Chikusei}. The reconstruction error analysis in Fig. \ref{fig:cave} highlights KANO’s strength in recovering both fine textures and coarse structures. For spectral fidelity, Fig. \ref{fig:pavia} shows that KANO produces spectral curves most closely aligned with the ground truth. This illustrates its advantage in modeling complex spectral distributions. Additionally, Fig. \ref{fig:Chikusei} demonstrates that KANO effectively reconstructs both the regular boundaries of agricultural regions and the irregular edges of water bodies, confirming its robustness in preserving spatial details.


\begin{figure*}[!t]
 	  \centering
 			\includegraphics[width=1.0\textwidth]{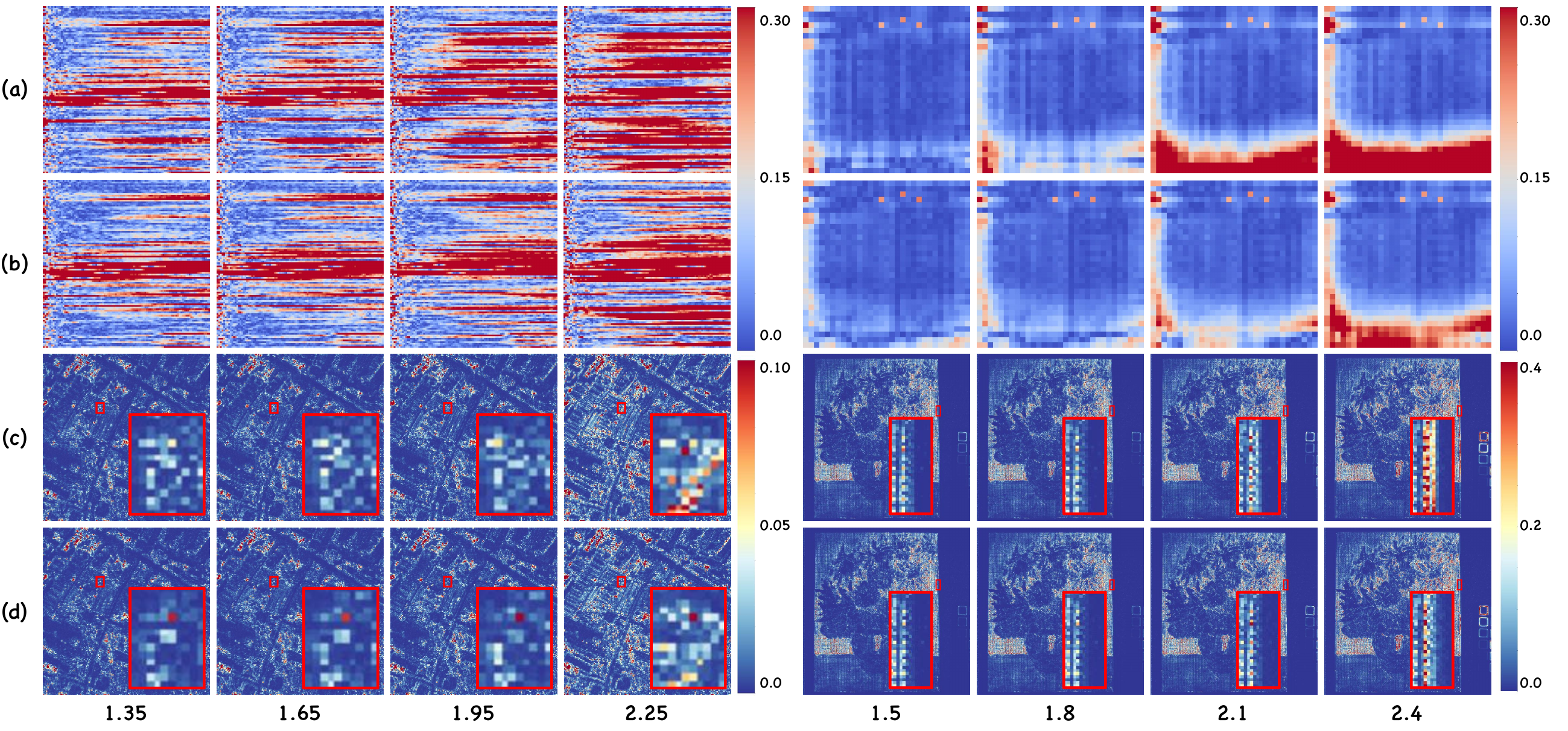}
          \caption{Spectral and spatial reconstruction errors of KANO with MLP and KAN backbones on Pavia and CAVE. (a),(b): Spectral RMSE per pixel-band; (c),(d): Spatial MSE maps. KAN-based KANO shows lower errors across both domains, especially as blur variance increases.}
  \label{figkanmlp_mse}
  \end{figure*}
\begin{table*}[!t]
    \centering
\caption{\textbf{Comparative analysis between \textcolor{RubineRed}{KANO} and \textcolor{cyan}{MLP} performance on diverse levels of complexity (spectral bands from low to high latitudes).} \textbf{Dep.} pertains to distinct experimental conditions involving the various network architectures (in terms of the number of layers). Note that the KAN module in KANO adopts the same layer configuration as in \textbf{Dep.2}. Throughout this paper, the best and second-best results of each test case are highlighted in \textcolor{RubineRed}{\textbf{Bold}}, \textcolor{cyan}{\underline{blue}}, respectively.}
\scalebox{1.0}{
    \begin{tabular}{c|c|cccc|cccc}
        \Xhline{1.0pt} 
         \cellcolor{gray!10}{}&\cellcolor{gray!10}{{\textbf{test data}}} & \multicolumn{4}{c|}{\cellcolor{gray!10}{\textbf{CAVE}}} & \multicolumn{4}{c}{\cellcolor{gray!10}{\textbf{Pavia}}}\\
          \hhline{>{\arrayrulecolor{gray!10}}->{\arrayrulecolor{black}}|-|-|-|-|-|-|-|-|-|}
             \multirow{-2}{*}{\textbf{\cellcolor{gray!10}{\textbf{Comparison Setting}}}}& \multicolumn{1}{c|}{\cellcolor{gray!10} {\textbf{index}}} & \cellcolor{gray!10} {$\sigma$=1.65} & \cellcolor{gray!10} {$\sigma$=1.8} & \cellcolor{gray!10} {$\sigma$=1.95} & \cellcolor{gray!10} {$\sigma$=2.1} & \cellcolor{gray!10} {$\sigma$=1.65} & \cellcolor{gray!10} {$\sigma$=1.8} & \cellcolor{gray!10} {$\sigma$=1.95} & \cellcolor{gray!10} {$\sigma$=2.1}\\
        \hline
        \hline
         & \textbf{Dep.1} &40.56 &40.42 &40.15 &39.53 &35.03 &34.79 &34.47 &33.90 \\
         & \textbf{Dep.2} &\textcolor{cyan}{\underline{40.97}} &\textcolor{cyan}{\underline{40.78}} &\textcolor{cyan}{\underline{40.54}} &\textcolor{cyan}{\underline{39.89}} &\textcolor{cyan}{\underline{35.21}} &\textcolor{cyan}{\underline{35.02}} &34.70 &34.07 \\
         & \textbf{Dep.3} &40.50 &40.37 &40.08 &39.45 &\textcolor{cyan}{\underline{35.21}} &35.01& 34.73& 34.14 \\
         \multirow{-4}{*}{\textbf{\textcolor{cyan}{MLP}}} & \textbf{Dep.4} & 39.74& 39.51& 39.10& 38.44 & 35.18& 35.00& \textcolor{cyan}{\underline{34.76}}& \textcolor{cyan}{\underline{34.30}} \\
        \hline
        \multicolumn{2}{c|}{\cellcolor{red!5}{\textcolor{RubineRed}{\textbf{KANO}}}} & \cellcolor{red!5}{\textcolor{RubineRed}{\textbf{41.08}}} & \cellcolor{red!5}{\textcolor{RubineRed}{\textbf{40.88}}} & \cellcolor{red!5}{\textcolor{RubineRed}{\textbf{40.59}}} & \cellcolor{red!5}{\textcolor{RubineRed}{\textbf{39.91}}} & \cellcolor{red!5}{\textcolor{RubineRed}{\textbf{35.29}}} & \cellcolor{red!5}{\textcolor{RubineRed}{\textbf{35.10}}} & \cellcolor{red!5}{\textcolor{RubineRed}{\textbf{34.84}}} & \cellcolor{red!5}{\textcolor{RubineRed}{\textbf{34.31}}} \\
        \Xhline{1.0pt}
    \end{tabular}}
    \label{tab:abl_multi_mlp}
\end{table*}

\subsection{Ablation Study} \label{ablation_studies}
KANO comprises three core components: layer-wise kernel approximation capability, a multi-scale fitting strategy, and a nonlinear compensation constraint module. This section presents a comprehensive analysis of the contribution of each component and investigates the comparative fitting performance of KAN and MLP on images of varying complexity.

\subsubsection{Approximation Capability under Different Depth} 
As shown in Tab. \ref{tab:abl_depth}, increasing the number of KAN layers leads to a performance trend that initially improves but then declines across all five natural image datasets. Notably, Set5 and Set14 exhibit more pronounced fluctuations, up to 0.2 dB, suggesting that model depth has a greater influence on datasets with higher internal consistency. Similarly, in hyperspectral image performance with varying KAN depths, the best results are obtained with relatively shallow KAN configurations, indicating that a small number of layers is sufficient to achieve strong modeling capacity. Beyond this point, additional depth introduces overfitting, leading to a decline in performance. 

\subsubsection{Multiscale Performance in KANO}  
To further evaluate the effectiveness of KAN in explicit modeling, we examine the influence of channel dimensionality during the fitting process. As shown in Tab. \ref{tab:abl_multi_mlp}, KANO exhibits a consistent increase-then-fall performance trend in all four data sets as channel scales vary. Additionally, comparing Tab. \ref{tab:abl_multi_mlp} reveals that incorporating channel scaling alongside layer depth variation leads to further performance gains. This indicates that adaptively adjusting channel dimensions enhances KANO's modeling capacity. Among the configurations, $\mathcal{C} \rightarrow 2\mathcal{C} \rightarrow \mathcal{C}$ achieves the highest PSNR and SSIM across all datasets.

\subsubsection{Nonlinear Module for Time-varying System}
As illustrated in Fig. \ref{fig:medifftmse} and \ref{fig:mlp_kernel}, the nonlinear compensation component $S$ exhibits the highest mean squared error (MSE), followed by the high-resolution estimate $O$, while the final super-resolved output $X$ achieves the lowest error. Furthermore, frequency-domain visualizations reveal that in early iterations, $S$ primarily contributes high-frequency details to $X$. As training progresses, the spectral distribution of $O$ increasingly approximates that of $X$, with the corrective influence of $S$ correspondingly diminishing. These findings underscore the pivotal role of $S$ in refining $O$, highlighting the effectiveness of the nonlinear fitting module in enhancing reconstruction fidelity within the KANO framework.

\subsection{Performance Comparison and Analysis between KANO and MLPs} 
To comprehensively evaluate the performance of KAN and MLP architectures, we conduct a comparative analysis under varying network depths and complexities. As shown in Fig.~\ref{fig:mlp_kernel}, the blur kernels predicted by MLP often exhibit sharp peaks but deviate from the ground truth (GT) in the overall structure. In contrast, KAN produces blur kernels with 2D heatmaps and 3D surface plots that more closely match the GT, demonstrating its superior ability to capture realistic kernel distributions. 

Tab.~\ref{tab:abl_multi_mlp} summarizes the reconstruction performance on the Pavia and CAVE datasets using either MLP or KAN for kernel estimation. The results indicate that KAN exhibits a clear advantage at shallower depths when dealing with kernels of larger variance, whereas deeper architectures perform better under smaller variances. On the Pavia dataset, the performance gain of KAN becomes more pronounced with increasing kernel variance. These trends may be attributed to dataset characteristics: CAVE, with fewer spectral bands, places greater importance on model depth, while Pavia’s high-dimensional spectral information challenges MLP's capacity to capture complex representations. KAN, by contrast, demonstrates a stronger ability to model such complexity, enabling superior reconstruction under severe degradation.

Fig.~\ref{figkanmlp_mse} further visualizes MSE distributions across spatial and spectral dimensions. On Pavia, MLP produces more widespread spatial errors, whereas on CAVE, it yields higher spectral errors, particularly for pixels with complex reflectance profiles. Fig.~\ref{figkanmlp_mse} (c) and (d) show that increasing the kernel variance degrades the performance for both methods. However, KAN consistently achieves lower MSEs across both datasets, confirming its superior capability in preserving fine-grained spatial and spectral structures, even under challenging conditions.

\section{Conclusion}\label{sec:conclusion}
In this paper, we applied the KAN to the image super-resolution task proposed by KANO to demonstrate its ability to approximate continuous spectral bands with arbitrary precision. Specifically, we achieved the discrete decomposition of continuous spectral functions through modal decomposition, where state transitions over time are governed by the Koopman equation. To fully approximate the high-resolution spectral function, KANO considers both the fitting of the linear system through the KAN module and the nonlinear control deviations, coupled via skip connections to produce the final super-resolved image. Comprehensive experiments on synthetic data across different spectral bands, along with ablation studies on the designed network, thoroughly validate the effectiveness and versatility of KANO in this approach.

\ifCLASSOPTIONcompsoc
\else
\fi


\bibliographystyle{ieeetr}
\bibliography{KNO_ref}

\end{document}